\documentclass{article} 
\usepackage{iclr2021_conference,times}
\usepackage{lineno}

\usepackage{amsmath,amsfonts,bm}

\def\eqref#1{equation~\ref{#1}}

\def\1{\bm{1}}

\DeclareMathAlphabet{\mathsfit}{\encodingdefault}{\sfdefault}{m}{sl}
\SetMathAlphabet{\mathsfit}{bold}{\encodingdefault}{\sfdefault}{bx}{n}

\usepackage{comment}
\usepackage{hyperref}
\usepackage{graphicx}
\usepackage{url}
\usepackage{multirow}
\usepackage{placeins}

\title{\center{When and how CNNs generalize to \\  out-of-distribution \\ category-viewpoint combinations}}

\author{Spandan Madan$^\dagger$ \\
  SEAS, Harvard University\\
   \And
  Timothy Henry \\
  CBMM \& MIT\\
   \\
  \And 
  Jamell Dozier \\
 CBMM \& MIT\\
  \And
  Helen Ho\\
  MIT CSAIL\\
  \And
  Nishchal Bhandari\\
  MIT CSAIL\\
  \And
  Tomotake Sasaki\\
  Fujitsu Laboratories Ltd.\\
  \And
  Fr\'edo Durand\\
  MIT CSAIL\\
  \And
  Hanspeter Pfister\\
  SEAS, Harvard University\\
  \And
  Xavier Boix$^\dagger$\\
  CBMM \& MIT\\
}

\iclrfinalcopy 
\begin{document}
\setlength{\textwidth}{14.25cm}

\maketitle

\newcommand\blfootnote[1]{%
  \begingroup
  \renewcommand\thefootnote{}\footnote{#1}%
  \addtocounter{footnote}{-1}%
  \endgroup
}

\newcommand\extrafootertext[1]{%
    \bgroup
    \renewcommand\thefootnote{\fnsymbol{footnote}}%
    \renewcommand\thempfootnote{\fnsymbol{mpfootnote}}%
    \footnotetext[0]{#1}%
    \egroup
}

\extrafootertext{$^\dagger$ \emph{Corresponding authors:} \texttt{spandan\_madan@g.harvard.edu} and \texttt{xboix@mit.edu}}

\newcommand*{\eg}{\emph{e.g.,~}}
\newcommand*{\ie}{\emph{ie.,~}}
\newcommand*{\etal}{\emph{et al.~}}
\newcommand*{\etc}{\emph{etc.}}

\begin{abstract}

  Object recognition and viewpoint estimation lie at the heart of visual understanding. Recent works suggest that convolutional neural networks (CNNs) fail to generalize to  out-of-distribution (OOD) category-viewpoint combinations, \ie combinations not seen during training.  In this paper, we investigate when and how such OOD generalization may be possible by evaluating CNNs trained to classify both object category and 3D viewpoint on OOD combinations, and identifying the neural mechanisms that facilitate such OOD generalization. We show that increasing the number of in-distribution combinations (\ie data diversity) substantially improves generalization to OOD combinations, even with the same amount of training data. We compare learning category and viewpoint in separate and shared network architectures, and observe starkly different trends on in-distribution and OOD combinations,~\ie while shared networks are helpful in-distribution, separate networks significantly outperform shared ones at OOD combinations.  Finally, we demonstrate that such OOD generalization is facilitated by the neural mechanism of specialization,~\ie the emergence of two types of neurons---neurons selective to category and invariant to viewpoint, and vice versa.

\end{abstract}

The combination of object recognition and viewpoint estimation is essential for effective visual understanding. 
In recent years, convolutional neural networks (CNNs) have offered state-of-the-art solutions for both these  fundamental tasks~\citep{he2016deep,szegedy2016rethinking,huang2017densely,su2015render,massa2016crafting, elhoseiny2016comparative, mahendran2018convolutional, afifi2018simultaneous}. However, recent works also suggest that CNNs have a hard time generalizing to combinations of object categories and viewpoints not seen during training, \ie out-of-distribution (OOD) generalization is a challenge. For object recognition, works have shown CNNs struggling to  generalize across spatial transformations like 2D rotation and translation~\citep{engstrom2017exploring,azulay2019deep,srivastava2019minimal}, and non-canonical 3D views~\citep{alcorn2019strike,barbu2019objectnet}. For viewpoint estimation, previous works propose learning category specific models~\citep{massa2016crafting, tulsiani2015viewpoints} or feed class predictions as input to the model~\citep{xiang2017posecnn, manhardt2020cps}, as generalizing to novel categories is a challenging task.

\begin{figure*}[t!]
\begin{tabular}{@{\hspace{0.1cm}}c@{\hspace{0.1cm}}c@{\hspace{0.1cm}}c}
\centering\includegraphics[width=0.32\linewidth]{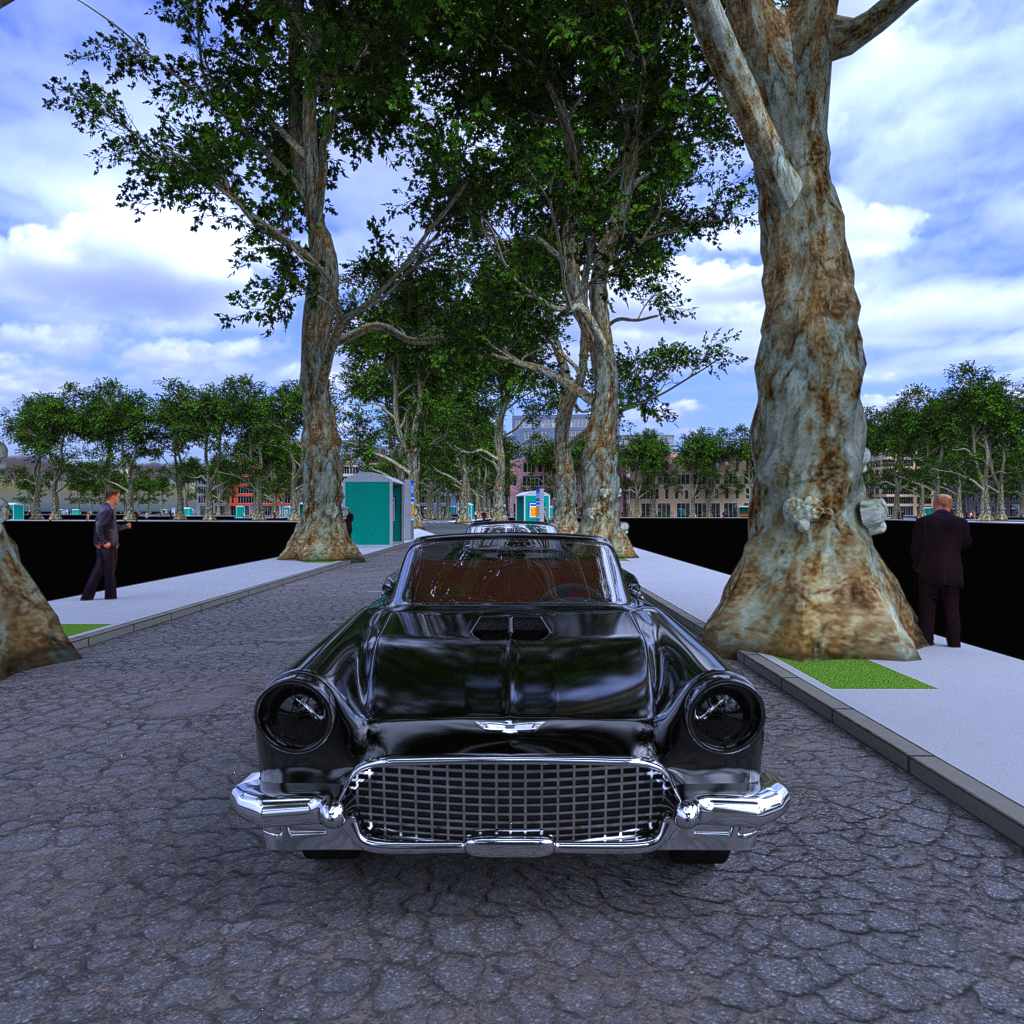}&
     \includegraphics[width=0.32\linewidth]{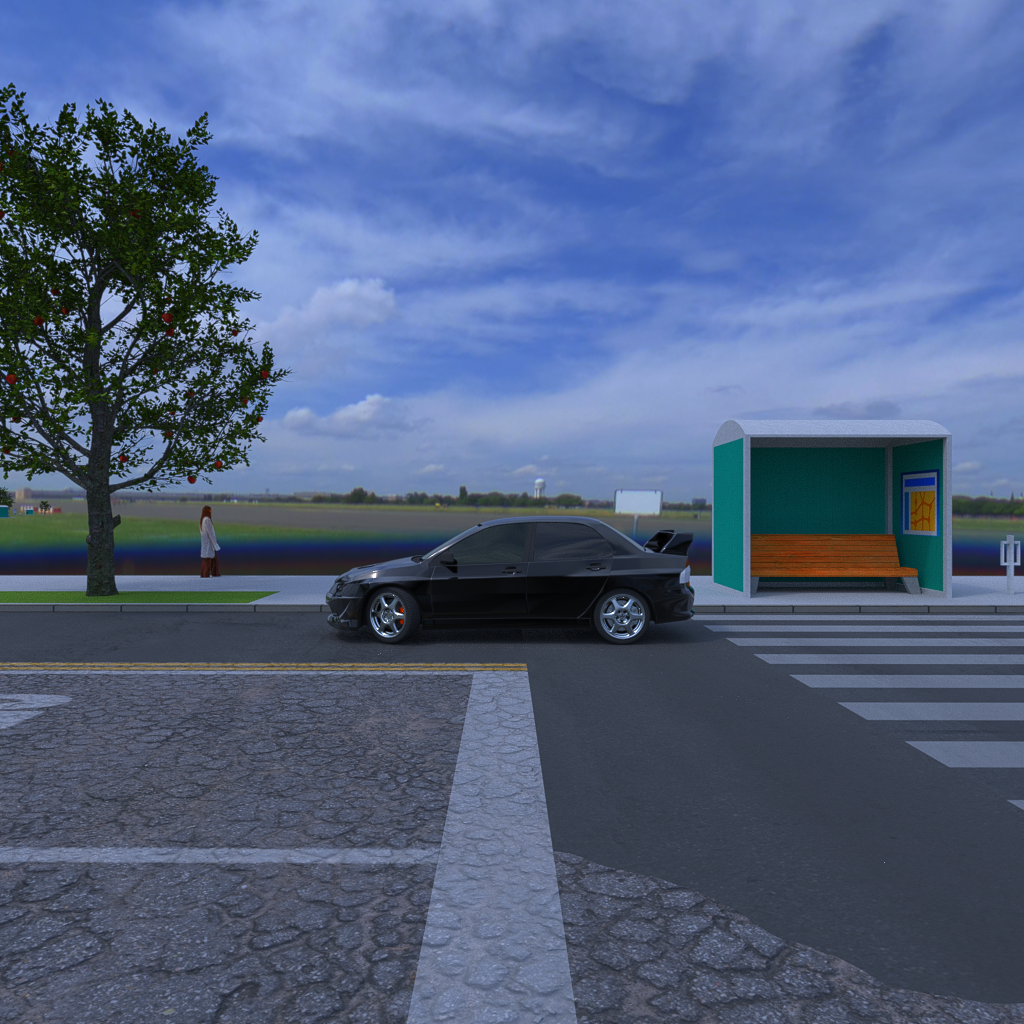}
& \includegraphics[width=0.32\linewidth]{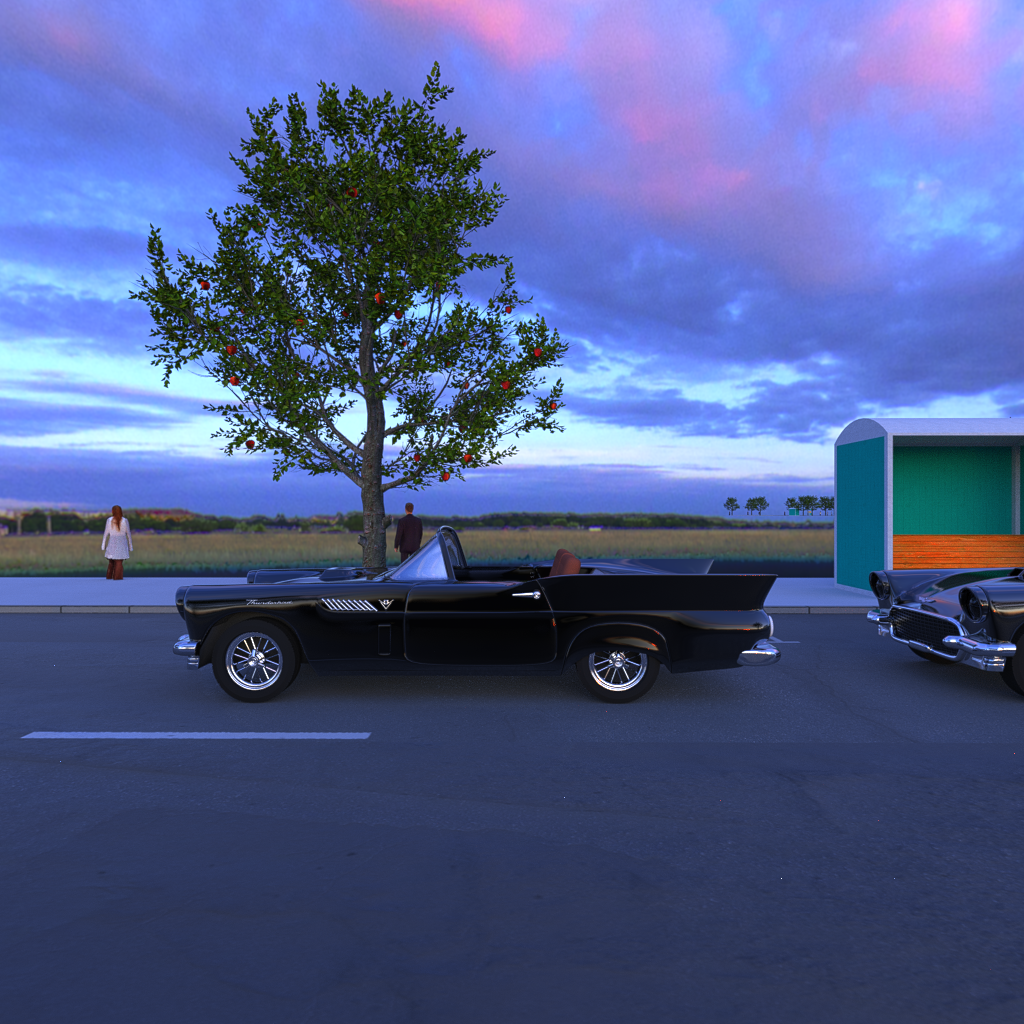}\\
\small{Ford Thunderbird, Front} & \small{Mitsubishi Lancer, Side} & \small{Ford Thunderbird, Side} \\
& (a) &\\[0.75em]
\centering\includegraphics[width=0.32\linewidth]{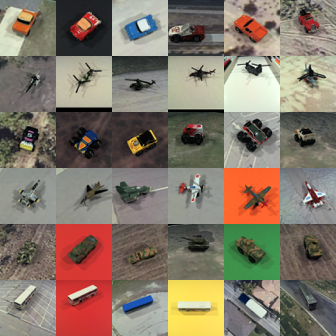}
& \includegraphics[width=0.32\linewidth]{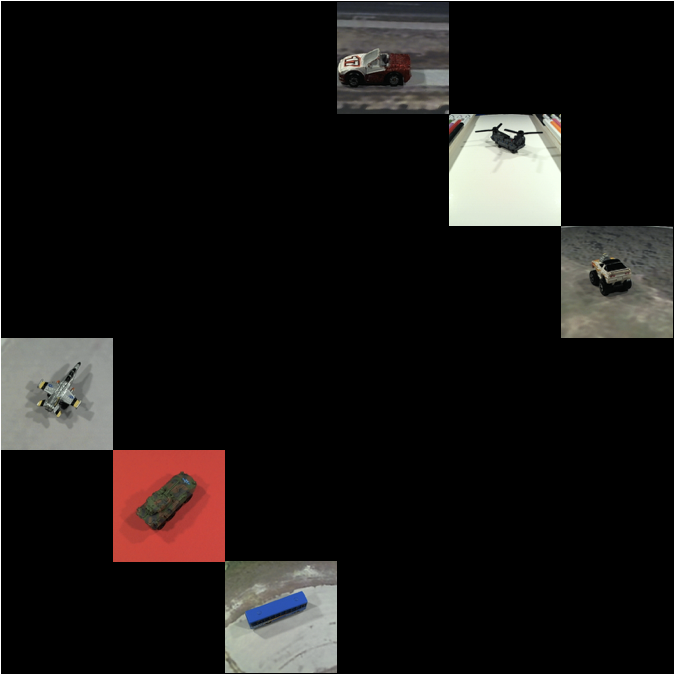}
&
     \includegraphics[width=0.32\linewidth]{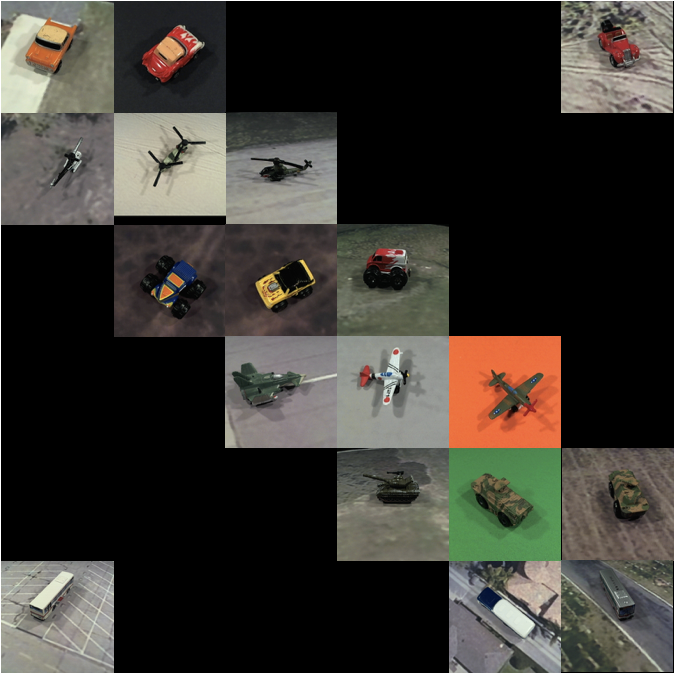}
     \\
\small{iLab-2M dataset}  &   \small{OOD Combinations (held-out)} & \small{ $50\%$ in-distribution Combinations } \\
(b) & (c) & (d)\\

\end{tabular}
\caption{\emph{Category-Viewpoint datasets.} (a) Our new \textit{Biased-Cars} dataset: Can a network shown only the Ford Thunderbird from front and the Mitsubishi Lancer from side generalize to classify the category and viewpoint for a Thunderbird seen from the side? (b) iLab-2M dataset~\citep{borji2016ilab}: Each cell represents a unique category-viewpoint combination (categories vary between rows, viewpoints between columns) with multiple object instances per category and backgrounds. (c) Held-out test set of category-viewpoint combinations. Same held-out test set is used to evaluate networks trained with different number of in-distribution combinations. (d) Biased training set with $50\%$ of category-viewpoint combinations. Number of categories and viewpoints selected is always equal. }
\label{multi_attribute_datasets}
\end{figure*}

It remains unclear \emph{when} and \emph{how} CNNs may generalize to OOD category-viewpoint combinations. Fig.~\ref{multi_attribute_datasets}a presents a motivating example: would a network trained on examples of a Ford Thunderbird seen only from the front, and a Mitsubishi Lancer seen only from the side generalize to predict car model (category) and viewpoint for a Thunderbird shown from the side? If so, what underlying mechanisms enable such OOD generalization?

In this paper, we investigate the impact of two key factors (data diversity and architectural choices) on the capability of generalizing to OOD combinations, and the neural mechanisms that facilitate such generalization. Concretely, we introduce the following discoveries:
\begin{enumerate}
    \item \textbf{Data diversity significantly improves OOD performance, but degrades in-distribution performance:} We investigate the role of data diversity by varying the number of in-distribution category-viewpoint combinations, keeping dataset size constant. We find that data diversity matters significantly. For a constant dataset size, increasing data diversity makes the task more challenging, as reflected in the deteriorating in-distribution performance. Yet, increasing data diversity substantially improves performance on OOD combinations. 
    
    \item \textbf{Separate architectures significantly outperform shared ones on OOD combinations unlike in-distribution:} We also analyze the performance of different architectures in the multi-task setting of simultaneous category and viewpoint classification,~\ie learning category and viewpoint in \emph{Shared} or in \emph{Separate} (no layers shared) architectures.
     Our results reveal that \emph{Separate} architectures generalize substantially better to OOD combinations compared to \emph{Shared} architectures. Also, this trend is in stark contrast with the trend for in-distribution combinations, where \emph{Shared} architectures perform marginally better. Thus, the belief that \emph{Shared} architectures outperform \emph{Separate} ones when tasks are synergistic should be revisited~\citep{caruana1997multitask}, as their relative performance strongly depends on whether the test sample is in-distribution or OOD.

    \item \textbf{Neural specialization facilitates generalization to OOD combinations:}
    Existing works suggest that OOD generalization is facilitated by selective and invariant representations~\citep{giles1987learning,riesenhuber1998just,goodfellow2009measuring,achille2018emergence}. However, this has not been previously demonstrated for deep learning, and does not extend to simultaneous category and viewpoint classification. To address this, we propose the neural mechanism of specialization---the emergence of two types of neurons, one driving OOD generalization for category, and the other for viewpoint. This corresponds to neurons selective to a category and invariant to viewpoint, and vice versa. We show that the CNN generalization behavior trends correlates with the degree of specialization of the neurons. 
\end{enumerate}

These results are consistent across multiple CNNs and datasets including the natural image dataset iLab-2M~\citep{borji2016ilab}, variations of MNIST~\citep{lecun1998gradient} extended with position and scale, and a challenging new dataset of car model recognition and viewpoint estimation---the \textit{Biased-Cars} dataset, which we introduce in this paper. This dataset consists of $15$K photo-realistic rendered images of several car models at different positions, scales and viewpoints, and under various illumination, background, clutter and occlusion conditions. With this, we hope to provide a first milestone in understanding the underlying mechanisms which enable OOD generalization in Multi-Task Learning for category and viewpoint classification.

\section*{Datasets for simultaneous category-viewpoint classification}\label{section:datasets}

Most existing datasets with category and viewpoint labels~\citep{xiang2014beyond, caesar2019nuscenes,barbu2019objectnet,min2019spair} present two major challenges - (i) lack of control over the distribution of categories and viewpoints, or (ii) small size. Thus, we present our results on the following datasets which do not suffer from these challenges:

\noindent {\bf iLab-2M dataset:}
iLab-2M~\citep{borji2016ilab} is a large scale (two million images), natural image dataset with $3$D variations in viewpoint and multiple object instances for each category (Fig.\ref{multi_attribute_datasets}b). The dataset was created by placing toy objects on a turntable and photographing them from six different azimuth viewpoints, each at five different zenith angles (total $30$). From the original dataset, we chose a subset of six object categories - Bus, Car, Helicopter, Monster Truck, Plane, and Tank. In Fig.~\ref{multi_attribute_datasets}b, each row represents images from one category, and each column images from one azimuth angle. All networks are trained to predict one of six category and viewpoint (azimuth) labels each.

\noindent {\bf MNIST-Position and MNIST-Scale:}
Inspired by the MNIST-Rotation dataset~\citep{larochelle2007empirical} which adds rotation to MNIST~\citep{lecun1998gradient} images, we created two more variants by adding viewpoint in the form of position or scale. MNIST-Position was created by placing MNIST images into one of nine possible locations in an empty 3-by-3 grid. For MNIST-Scale we resized images to one of nine possible sizes followed by zero-padding. Images of the digit $9$ were left out in both these datasets, ensuring nine categories and nine viewpoints classes (total of $81$ category-viewpoint combinations). Sample images are available in the supplement~\ref{sec:supMNIST}.

\noindent {\bf Biased-Cars dataset:} Building on other multi-view car datasets for viewpoint estimation~\citep{KrauseStarkDengFei-Fei_3DRR2013,ozuysal2009pose}, we introduce a challenging new dataset for simultaneous object category and viewpoint classification---the \textit{Biased-Cars} dataset. Our dataset features photo-realistic outdoor scene data with fine control over scene clutter (trees, street furniture, and pedestrians), car colors, object occlusions, diverse backgrounds (building/road textures) and lighting conditions (sky maps). \textit{Biased-Cars} consists of $15$K images of five different car models seen from viewpoints varying between $0$-$90$ degrees of azimuth, and $0$-$50$ degrees of zenith across multiple scales. Our dataset offers two main advantages: \emph{(a)} complete control over the joint distribution of categories, viewpoints, and other scene parameters, and \emph{(b)} unlike most existing synthetic city datasets~\citep{qiu2016unrealcv,caesar2019nuscenes,dosovitskiy2017carla} we use physically based rendering for greater photo-realism, which has been shown to help networks transfer to natural image data significantly better~\citep{zhang2017physically,halder2019physics}. Sample images are shown in Fig.~\ref{multi_attribute_datasets}a. As in~\citep{xiang2014beyond,divon2018viewpoint}, we choose to focus on azimuth prediction. The azimuth is divided into five bins of $18$ degrees each, thus ensuring five category (car models) and five viewpoint classes (azimuth bins), for a total of $25$ different category-viewpoint combinations.

\textbf{Additional Datasets:} In the supplement we provide results on two additional standard datasets---MNIST-Rotation~\citep{larochelle2007empirical} and the UIUC3D dataset~\citep{savarese20073d}. Note that the UIUC dataset has a skewed joint distribution of category-viewpoint combinations. This makes it difficult to run controlled experiments. However, the experiments which were possible on this dataset confirm that our findings extend to it as well. 

For all datasets, networks are trained to classify both category and viewpoint simultaneously without pretraining, and the number of classes for each task is kept equal to ensure equal treatment. More details can be found in  supplement~\ref{sec:suppCars}. As shown in the experiments, these datasets are challenging benchmarks for testing generalization, with a huge scope for improvement for state-of-the-art CNNs.

\section*{Factors affecting generalization behaviour}

Below we present the two factors we study for their impact on generalization to OOD category-viewpoint combinations - \emph{(i)} data diversity, and \emph{(ii)} architectural choices.

\noindent \paragraph{Generating train/test splits with desired data diversity.}\label{sec:hold_out} All our datasets can be visualized as a square category-viewpoint \emph{combinations grid} as shown for the iLab dataset in Fig.~\ref{multi_attribute_datasets}b. Each row represents images from one category, and each column a viewpoint, \ie each cell represents all images from one category-viewpoint combination. 

For each dataset, we start by constructing an OOD test split---a set of category-viewpoint combinations are selected and held out from the \textit{combinations grid} as shown in Fig.~\ref{multi_attribute_datasets}c. We refer to these as the OOD combinations. Images from OOD combinations are never shown to any network during training. These images are only used to evaluate how networks generalize outside the training distribution. For a fair representation of each category and viewpoint, we ensure that every category and viewpoint class occurs exactly once in the OOD combinations,~\ie one cell each per row and column is selected.

Remaining cells in the combinations grid are used to construct multiple training splits with an increasing number of category-viewpoint combinations \ie data diversity. For each training split, we first sample a set of combinations as shown in Fig.~\ref{multi_attribute_datasets}d, which we call the in-distribution combinations. Then, we build the training data-split by sampling images from these in-distribution combinations. We ensure that every category and viewpoint occurs equally in the in-distribution combinations, \ie equal number of cells per each row and column. Fig.~\ref{multi_attribute_datasets}d shows the 50\% in-distribution training split for the iLab dataset. To ensure that we evaluate the effect of data diversity and not that of data amount, the number of images is kept constant across train splits as the number of in-distribution combinations is increased. Thus, the number of images per combination decreases as the number of in-distribution combinations is increased. Also, note that every network is trained with only one of these training splits at a time,~\ie data diversity is kept constant during training.

\noindent \paragraph{Architectural choices.}\label{architectures_sec}
One central question addressed in this paper is the impact of architectural choices on the capability to generalize to OOD category-viewpoint combinations. While many separate models have been proposed for object recognition and viewpoint estimation~\citep{ghodrati20142d, tulsiani2015pose}, recent years have seen a growing a trend of multi-task learning inspired architectures which suggest that recognition models can benefit from an understanding of object viewpoint, and vice versa~\citep{penedones2012improving, zhao2017improved, massa2016crafting,su2015render,li2018unified}. These architectures often learn a shared representation for both tasks, followed by task specific branches~\citep{su2015render, zhao2017improved,grabner20183d}.\\

\begin{figure*}[t!]
\begin{tabular}{@{\hspace{0.1cm}}c@{\hspace{0.1cm}}c@{\hspace{0.1cm}}c}
\centering\includegraphics[width=0.32\linewidth]{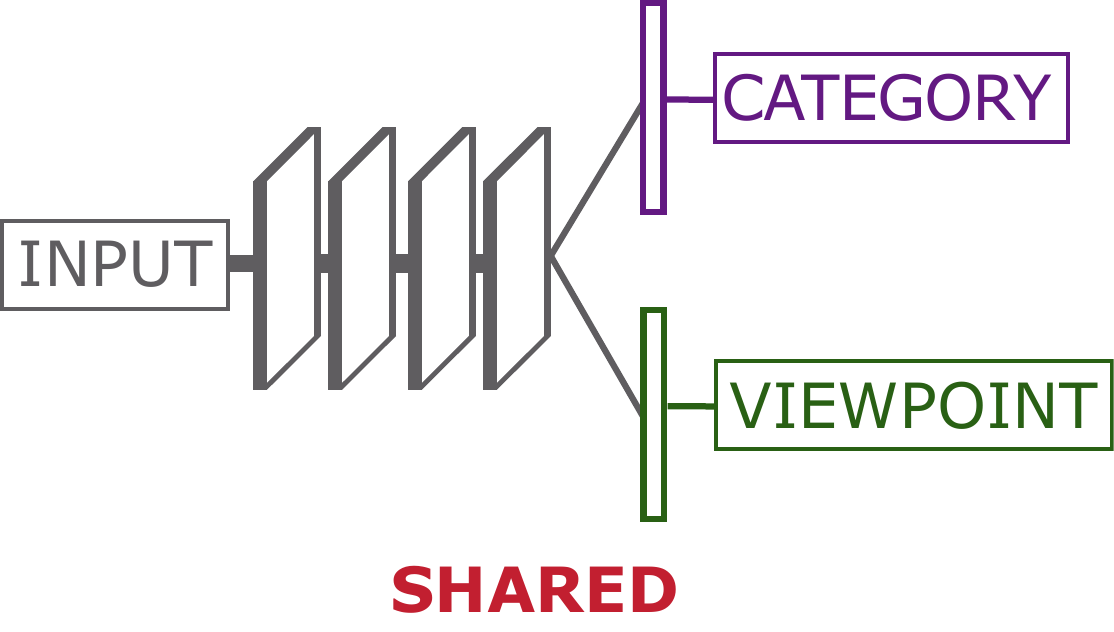}&
     \includegraphics[width=0.32\linewidth]{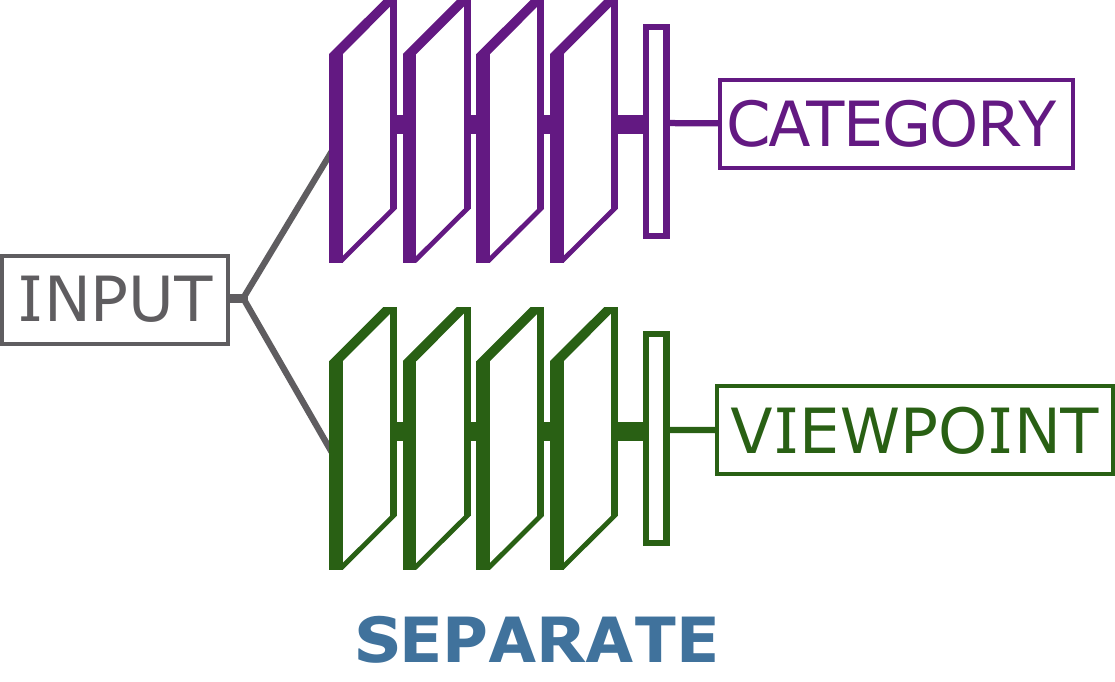}
& \includegraphics[width=0.32\linewidth]{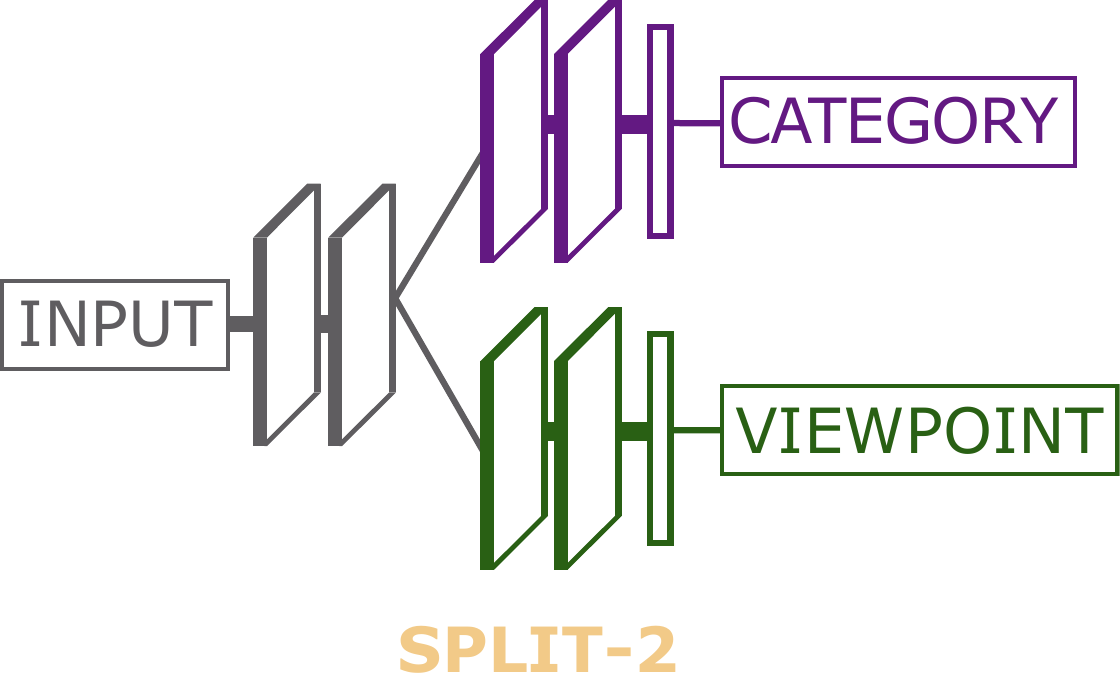}\\

\end{tabular}
 \caption{\emph{Architectures for Category Recognition and Viewpoint Estimation.} \textit{Shared}, \textit{Separate} and \textit{Split-2} architectures for ResNet-18. In the \textit{Shared} architecture, all layers until the last convolutional block are shared between tasks, followed by task specific fully connected branches. In the \textit{Separate} architecture, each task is trained in a separate network with no layer sharing. \textit{Split-2} presents a middle ground. These architectures are designed similarly for backbones other than ResNet-18.}
 \label{fig:architectures}
\end{figure*}

Here, we investigate the impact of learning shared representations on the network's capability to generalize to OOD category-viewpoint combinations \ie to extrapolate in the multi-task setting of simultaneous category and viewpoint classification. For this, we defined two types of backbone agnostic architectures---the \textit{Shared} and the \textit{Separate} architectures. Fig.~\ref{fig:architectures} depicts these architectures for a ResNet-18 backbone~\citep{he2016deep}. In the \textit{Shared} case, all convolutional blocks are shared between tasks, followed by task-specific fully connected layers, while there are no layers shared between tasks in the \textit{Separate} architecture. We also investigated $3$ additional \textit{Split} architectures which represent a gradual transition from \textit{Separate} to \textit{Shared} ResNet-18: the \textit{Split-1}, \textit{Split-2}, and \textit{Split-3} architectures. These were constructed by branching ResNet-18 after $1$, $2$, and $3$ convolutional blocks as shown in Fig.~\ref{fig:architectures}. Note that splitting at a layer leads to doubling of the number of neurons in that layer. In our experiments, we show that this increase in width does not provide an advantage.

\section*{Generalization through selectivity and invariance}\label{section:sel_invar}

Selectivity and invariance of neurons have long been hypothesized to facilitate generalization in both biological and artificial neural networks~\citep{bricolo19973d,riesenhuber1998just,goodfellow2009measuring,achille2018emergence,poggio2016visual,olshausen1993neurobiological,quiroga2005invariant,rust2010selectivity}.  Neurons are commonly interpreted as image feature detectors, such that the neuron's activity is high only when certain features are present in the image~\citep{zeiler2014visualizing,simonyan2013deep,zhou2014object,bau2017network,oquab2015object}. We refer to this property as \emph{selectivity} to an image feature. Selectivity alone, however, is not sufficient to generalize to OOD category-viewpoint combinations. For example, a neuron may be selective to features relevant to a category, but only so for a subset of all the viewpoints. Generalization is facilitated by selective neurons that are also \emph{invariant} to nuisance features. For instance, in Fig.~\ref{multi_attribute_datasets}a, neurons that are selective to the Ford Thunderbird and invariant to viewpoint would have very similar activity for the Ford Thunderbird on in-distribution and OOD viewpoints, thus enabling generalization to category recognition. Similarly, generalization to viewpoint estimation can be enabled by neurons selective to viewpoint and invariant to category.

Here, we present our implementation for quantifying the amount of \textit{selectivity} and \textit{invariance} of an individual neuron. Let $N$ be the number of categories or viewpoints in the dataset. We represent the activations for a neuron across all category-viewpoint combinations as an $N \times N$ \textit{activations grid}, as shown in Fig.~\ref{fig:specialization}a. Each cell in this \emph{activations grid} represents the average activation of a neuron for images from one category-viewpoint combination, with rows and columns representing average activations for all images from a single category (\eg Ford Thunderbird) and a viewpoint (\eg front), respectively. These activations are normalized to lie between $0$ and $1$ (see supplement~\ref{sec:supl:Norm}). For neuron $k$, we define $a_{ij}^k$ as the entry in the \textit{activations grid} for row (category) $i$ and column (viewpoint) $j$. Below we introduce the evaluation of a neuron's \textit{selectivity score} with respect to category and \textit{invariance score} with respect to viewpoint. Viewpoint selectivity score and category invariance score can be derived analogously.

\begin{linenomath*}
{\bf Selectivity score.} We first identify the category that the neuron is activated for the most on average,~\ie the category which has the maximum sum across the rows in Fig.~\ref{fig:specialization}a. We call this category the neuron's \textit{preferred category}, and denote it as $i^{\star k}$, such that $i^{\star k} = {\arg \max_i} \sum_{j} a_{ij}^k$. The selectivity score compares the average activity for the \emph{preferred category} (denoted as $\hat{a}^k$) with the average activity of the remaining categories ($\bar{a}^k$). Let $S^k_c$ be the selectivity score with respect to category, which we define as is usual in the literature (\eg~\cite{morcos2018importance,zhou2018revisiting}) with the following expression:   
\begin{equation}
     S^k_c = \frac{\hat{a}^k - \bar{a}^k}{\hat{a}^k + \bar{a}^k},
     \; \;  \; \; \mbox{where}  \;  \hat{a}^k = \frac{1}{N}\sum_{j} a_{i^{\star k} j}^k ,\;\; \bar{a}^k = \frac{\sum_{i\neq i^{\star k}}\sum_j {a}_{ij}^k}{N(N-1)}.\label{eqn:selectivity}
 \end{equation}
Observe that $S^k_c$ is a value between $0$ and $1$, and higher values of $S^k_c$ indicate that the neuron is more active for the \emph{preferred category} as compared to the rest. Selectivity with respect to viewpoint, denoted as $S^k_v$, can be derived analogously by swapping indices $(i,j)$.
\end{linenomath*}

\begin{linenomath*}
\noindent{\bf Invariance score.} A neuron's invariance to viewpoint captures the range of its average activity for the \textit{preferred category} as the viewpoint (nuisance parameter) is changed. Let $I^k_v$ be the invariance score with respect to viewpoint which we define as the difference between 
the highest and lowest activity across all viewpoints for the \emph{preferred category},~\ie
\begin{equation}
    I^k_v = 1 -\Big(\underset{j}{\max} \;\; a_{i^{\star k}j}^k - \underset{j}{\min} \;\; a_{i^{\star k}j}^k\Big),
\end{equation}
where the range is subtracted from $1$ to have the invariance score equal to $1$ when there is maximal invariance. Invariance with respect to category, denoted $I^k_c$, can be derived analogously.
\end{linenomath*}

\noindent {\bf Specialization score.} \label{subsection:specialization}Generalization to category recognition may be facilitated by neurons selective to category and invariant to viewpoint. Similarly, viewpoint selective and category invariant neurons can help generalize well to viewpoint estimation. This reveals a tension when category and viewpoint are learned together, as a neuron which is selective to category, cannot be invariant to category. The same is true for viewpoint. One way this contradiction may be resolved is the emergence of two types of neurons---category selective and viewpoint invariant, and vice versa. We refer to this as specialization. This hypothesis is well-aligned with the findings in~\citep{yang2019task}, which showed the emergence of groups of neurons contributing exclusively to single tasks. Thus, in the context of category recognition and viewpoint estimation, we hypothesize that neurons become selective to either category or viewpoint at later layers as the relevant image features for these tasks are disjoint (the category of an object cannot predict its viewpoint, and vice-versa).

\begin{linenomath*}
To classify neuron $k$ as a category or viewpoint neuron, we compare its selectivity for both category and viewpoint ($S^k_c$ and $S^k_v$). If  $S_c^k$ is greater than $S_v^k$, then neuron $k$ is a category neuron, otherwise, it is a viewpoint neuron. Since generalization capability relies on both invariance and selectivity, we introduce a new metric for a neuron, the specialization score denoted as $\Gamma^k$, which is the geometric mean of its selectivity and invariance scores,~\ie
\begin{equation}\label{eq:category_pose_neuron}
\Gamma^k = \left\{
        \begin{array}{ll}
            \sqrt{S^k_c I^k_v} & \mbox{if} \; S_c^k > S_v^k \quad (\text{category neuron}) \\
            \sqrt{S^k_v I^k_c} & \mbox{if} \; S_c^k \le S_v^k \quad (\text{viewpoint neuron})
        \end{array}
    \right. .
\end{equation}
Below, we present results that show that the specialization score is highly indicative of a network's performance on OOD combinations. 
\end{linenomath*}

\section*{When do CNNs generalize to OOD combinations?}\label{section:when}

\begin{figure*}[t!]
\begin{tabular}{@{\hspace{-0.cm}}c@{\hspace{-0.1cm}}c@{\hspace{-0.1cm}}c@{\hspace{-0.1cm}}c}
     \includegraphics[width=0.255\linewidth]{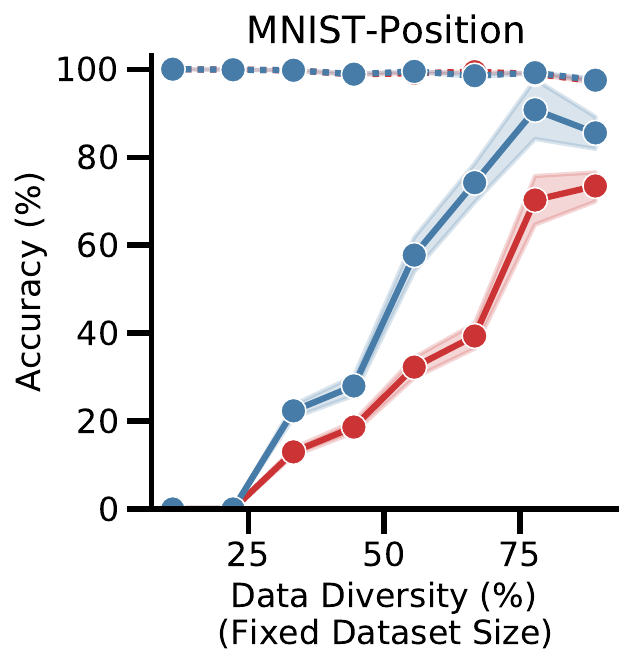}
&
     \includegraphics[width=0.255\linewidth]{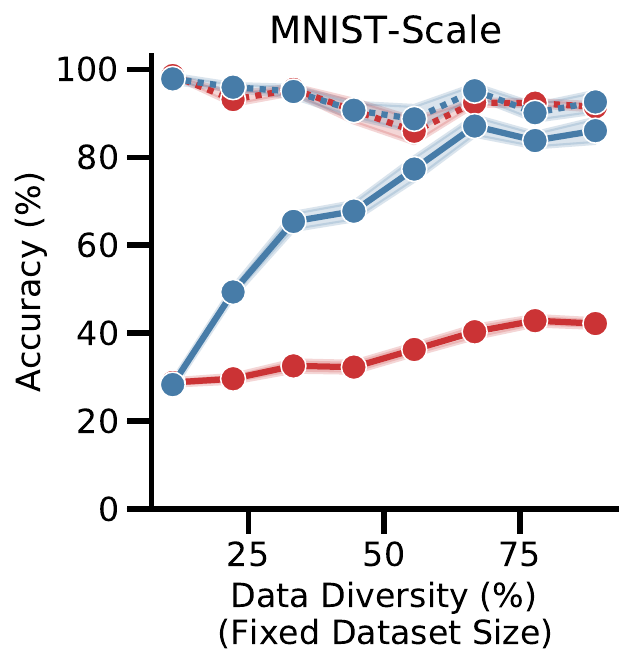}
&  \includegraphics[width=0.255\linewidth]{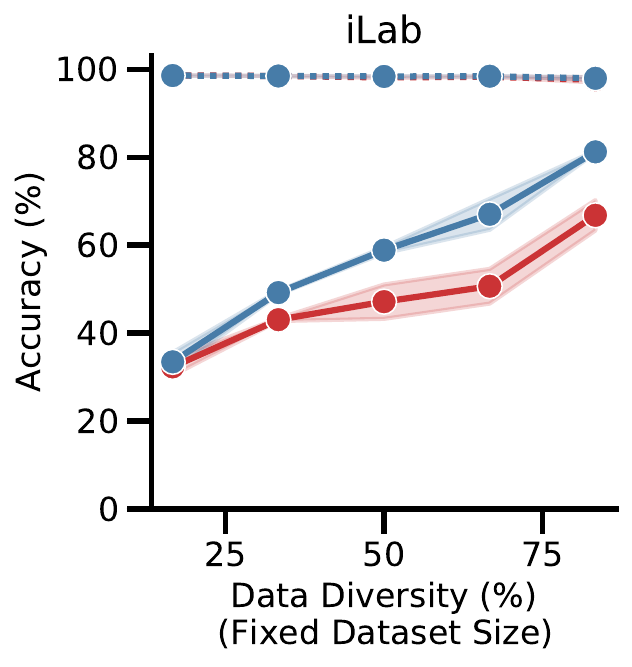} &  \includegraphics[width=0.255\linewidth]{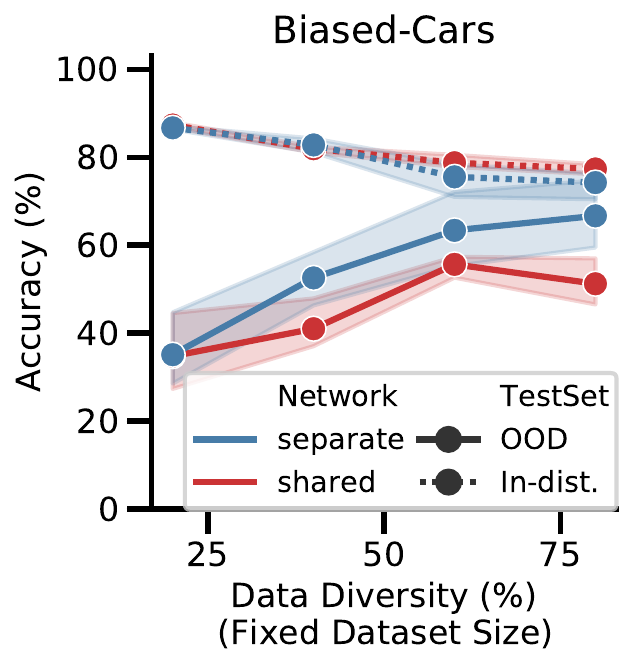} \\
(a) & (b) & (c)& (d) \\
\end{tabular}
\caption{\emph{Generalization performance for \textit{Shared} and \textit{Separate} ResNet-18 as in-distribution combinations are increased for all datasets.} The geometric mean of category recognition accuracy and viewpoint estimation accuracy is reported along with confidence intervals (95\%) (a)~MNIST-Position dataset. (b)~MNIST-Scale dataset. (c)~iLab dataset. (d)~Biased-Cars dataset.}
\label{fig:generalSation_datasets}
\end{figure*}

\begin{figure*}[t!]
\begin{tabular}{@{\hspace{-0.cm}}c@{\hspace{-0.1cm}}c@{\hspace{-0.1cm}}c@{\hspace{-0.1cm}}c}
\includegraphics[width=0.255\linewidth]{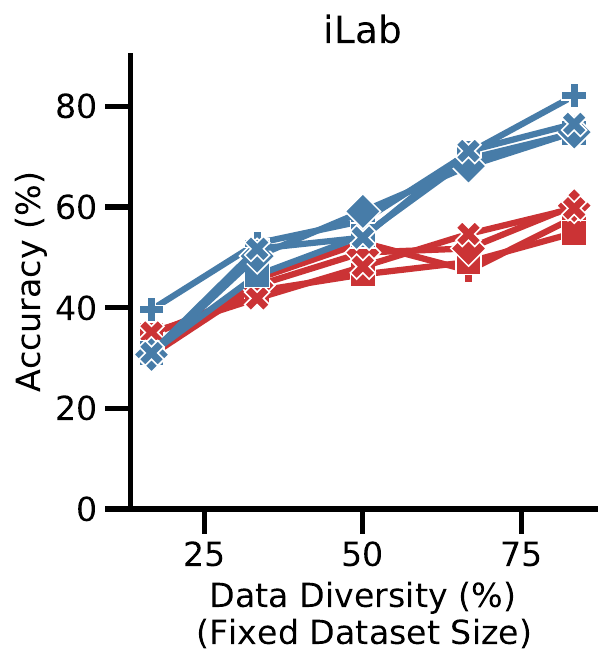} &  \includegraphics[width=0.255\linewidth]{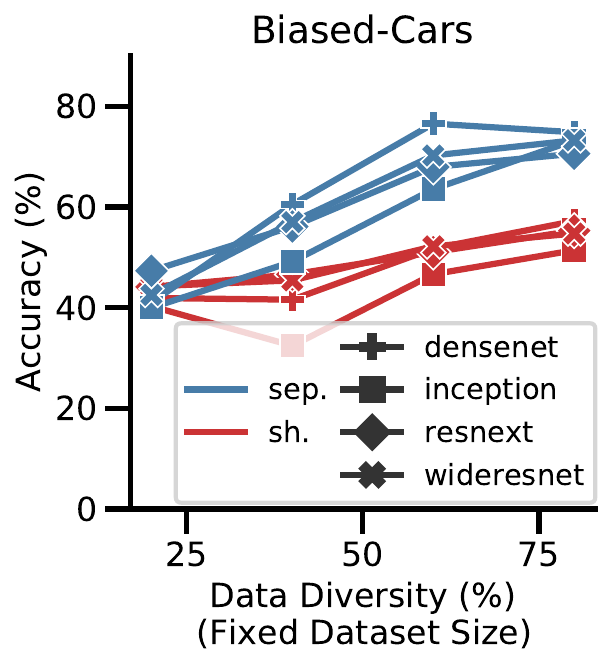}  
&
\includegraphics[width=0.255\linewidth]{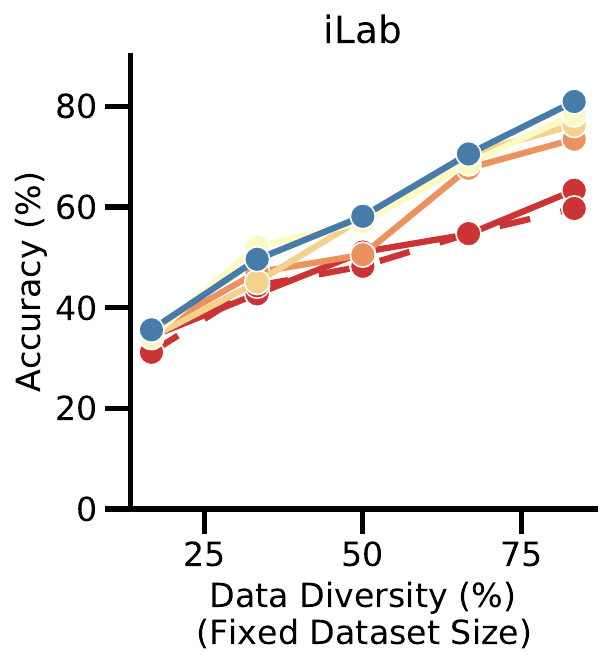}
&
     \includegraphics[width=0.255\linewidth]{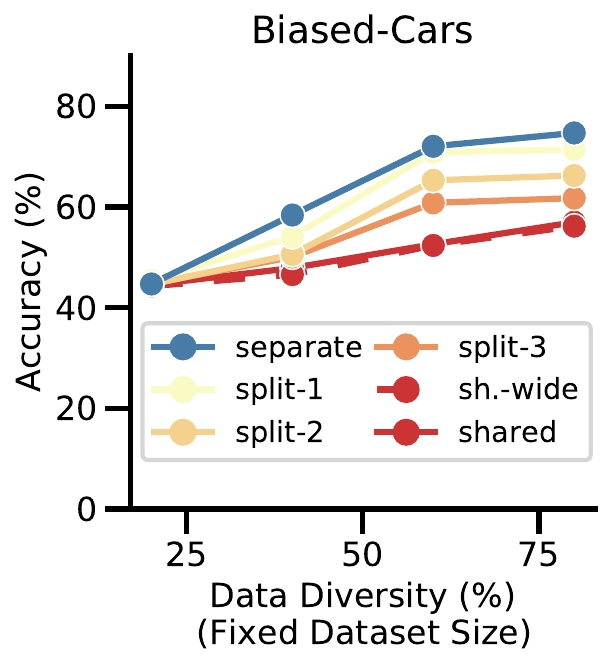}
 \\
(a) & (b) & (c)& (d) \\
\end{tabular}
\caption{\emph{Generalization performance for different architectures and backbones as in-distribution combinations are increased for iLab and Biased-Cars datasets.} The geometric mean between  category recognition accuracy and viewpoint recognition accuracy is reported for OOD combinations as number of in-distribution combinations is increased. 
(a) and (b) Accuracy of  \emph{Separate} and \emph{Shared} for backbones other than ResNet-18, for iLab and Biased-Cars datasets, respectively. 
(c) and (d) Accuracy of ResNet-18 \emph{Separate}, \emph{Shared} and different \textit{Split} architectures made by splitting at different blocks of the network, for iLab and Biased-Cars datasets, respectively. }
\label{fig:controls}
\end{figure*}


Below, we summarize our findings from evaluating \textit{Separate} and \textit{Shared} architectures when tested on unseen images from in-distribution and OOD category-viewpoint combinations. See supplement~\ref{sec:suplDetails} for experimental details.

\noindent {\bf For fixed dataset size, data diversity enables better OOD generalization, but deteriorates in-distribution performance.} Fig.~\ref{fig:generalSation_datasets} presents the geometric mean of category and viewpoint classification accuracy for \textit{Separate} and \textit{Shared} architectures with the ResNet-18 backbone, for all datasets. These experiments were repeated three times, and here we present the mean performance with confidence intervals. 
For fixed dataset size, increasing in-distribution combinations makes the task more challenging as images with each category and viewpoint become more diverse, leading to some drop in accuracy on in-distribution combinations. In contrast, both architectures show a significant improvement of their performance on images from OOD combinations, as data diversity increases. We ensured that this result can not be attributed to having closer viewpoint angles between in-distribution and OOD combinations as data diversity is increased (supplement~\ref{sec:suplSimilarity}). CNNs do not theoretically guarantee viewpoint invariance~\citep{poggio2016visual}, but our result provides reassurance that CNNs can become robust to OOD category-viewpoint combinations as long as they are shown enough diversity during training. Taken together, these results suggest an inherent trade-off between getting better on in-distribution combinations and extrapolating to OOD combinations, which is impacted by training data diversity. Also, these results add to a growing body of works investigating the trade-offs inherent to multi-task learning~\citep{standley2019tasks, shin2018pixels}.

Even though the geometric mean of category and viewpoint classification increases consistently with increased in-distribution combinations, individual accuracy for these tasks does not always increase consistently (see supplement~\ref{sec:suplSeparateAcc}). We attribute this to the randomness in the selection of in-distribution and OOD combinations. Furthermore, the relative accuracy of the two tasks varies depending on the dataset, and no task is consistently harder than the other across all datasets.

\noindent {\bf \emph{Separate} architectures generalize significantly better than \emph{Shared} ones in OOD combinations, unlike in-distribution.} A striking finding that emerged from our analysis is the contrast in the trends of the in-distribution and OOD performance. While both architectures perform well on new images from in-distribution combinations, \textit{Separate} architectures outperform \textit{Shared} ones by a very large margin on OOD combinations. For the ResNet-18 backbone, this result can be seen consistently across all $4$ datasets as shown in Fig.~\ref{fig:generalSation_datasets}. Supplement~\ref{sec:suplSeparateAcc} shows that \emph{Separate} also outperforms \emph{Shared} for category and viewpoint classification individually. Note that previous works have shown that \textit{Shared} architectures are superior for synergistic tasks, as networks can share features among tasks. These works test on the same combinations as seen during training (in-distribution), and when we do so, we also observe that \textit{Shared} architectures perform same or slightly better than \textit{Separate} ones (Fig.~\ref{fig:generalSation_datasets} dashed lines). Thus, our results reveal that the relative performance between \textit{Shared} and \textit{Separate} depends not only on the synergy between tasks, but also whether the evaluation is in-distribution or OOD.

We extended our analysis to \emph{Separate} and \emph{Shared} architectures with different backbones, namely ResNeXt~\citep{xie2017aggregated}, WideResNet~\citep{zagoruyko2016wide}, Inception v3~\citep{szegedy2016rethinking} and the DenseNet~\citep{huang2017densely}, as shown in Fig.~\ref{fig:controls}a and b. As can be seen, \emph{Separate} architectures outperform \emph{Shared} ones by a large margin for all backbones, which confirms that this result is not backbone specific. Investigating further, we experiment with \emph{Split} architectures, and as can be seen in Fig.~\ref{fig:controls}c and d, there is a consistent, gradual dip in the performance as we move from the \textit{Separate} to the \textit{Shared} architectures. Thus, generalization to OOD category-viewpoint combinations is best achieved by learning both tasks separately, with a consistent decrease in generalization as more parameter sharing is enforced.

To make sure that \textit{Separate} architectures do not perform better due to the added number of neurons, we made the \textit{Shared-Wide} architecture by doubling the neurons in each layer of the \textit{Shared} ResNet-18 network. As Fig.~\ref{fig:controls}c and d show, this architecture performs very similarly to the \textit{Shared} one (see additional results in ~\ref{app:NumberNeurons}). This is in accordance with previous results that show that modern CNNs may improve in performance as the width is increased but to a limited extent~\citep{nakkiran2019deep,casper2019removable}. 

In the supplement, we provide a number of additional controls that support the generality of our results. Concretely, we show results for different number of training images (supplement~\ref{app:NumberTraining}), viewpoint estimation for 4 new car models and category prediction for new viewpoints (supplement ~\ref{app:DifferentBias}), and the order in which category and viewpoint are learned (supplement ~\ref{app:TaskOrder}). We also present results on additional datasets (supplement~\ref{app:AdditionalDatasets}) and architectures (supplement~\ref{app:AdditionalArch}).

\section*{How do CNNs generalize to OOD combinations?}\label{section:how}

We now analyze the role of specialized (\ie selective and invariant) neurons in driving generalization to OOD category-viewpoint combinations.

\begin{figure*}[t!]
\begin{tabular}{@{\hspace{0.2cm}}c@{\hspace{-0.cm}}c@{\hspace{-0.1cm}}c}
\centering
 \multirow{1}{*}{
\begin{tabular}{c}
 \vspace*{-3.8cm}\\
\includegraphics[width=0.42\linewidth]{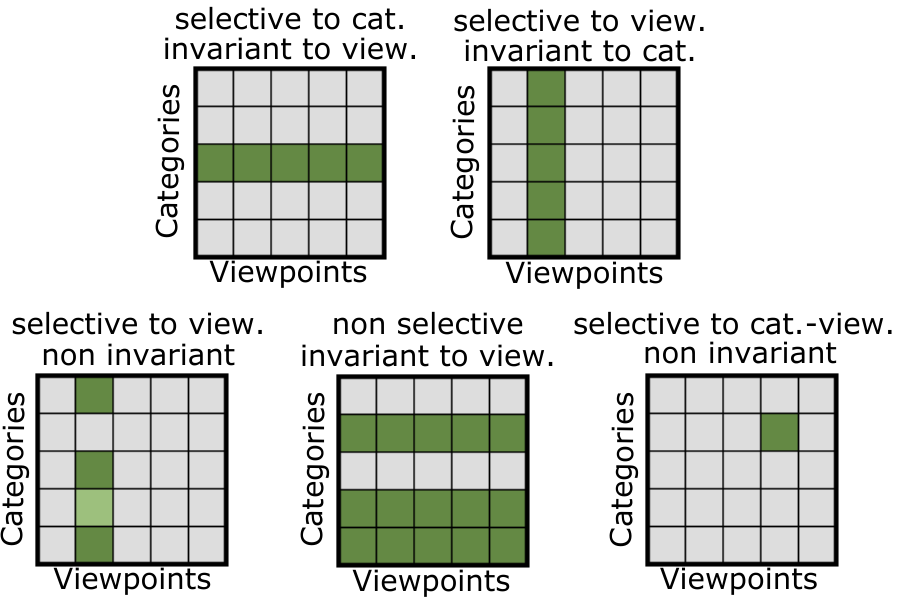}\\
(a)
\end{tabular}
}
&  \includegraphics[width=0.25\linewidth]{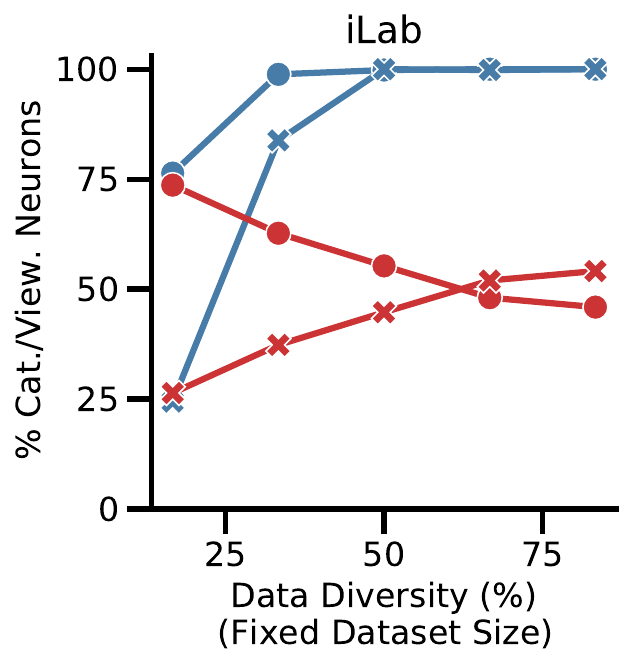} &  \includegraphics[width=0.25\linewidth]{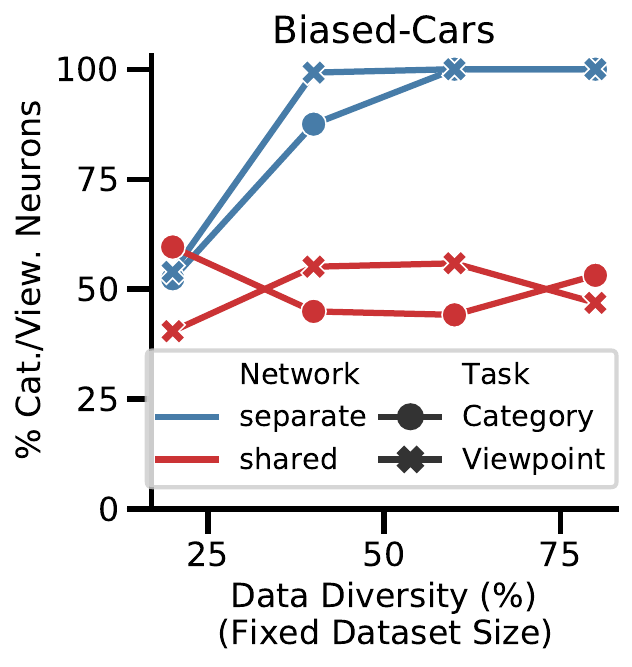}
 \\
 & (b) & (c) \\
\end{tabular}
\caption{\emph{Specialization to category recognition and viewpoint estimation.}  (a) Prototypical \emph{activation grids} for different types of selective and invariant neurons. (b) and (c) Percentage of neurons after ResNet-18 block-4 that are  specialized to category and viewpoint, for iLab and Biased-Cars datasets, respectively. ResNet-18 \emph{Separate} and \emph{Shared} networks are evaluated; for \emph{Separate}, only the task-relevant neurons for each branch are displayed.  \vspace{-0.4cm}}
\label{fig:specialization}
\end{figure*}

\noindent {\bf Specialization score correlates with generalization to OOD category-viewpoint.}
We first investigate the emergence of category and viewpoint neurons in the final convolutional layer of the networks. Fig.~\ref{fig:specialization}b and c show the percentage of neurons of each type in \emph{Shared} and \emph{Separate} architectures as in-distribution combinations are increased. As can be seen, all neurons in the category and viewpoint branches of the \textit{Separate} architecture become specialized to category and viewpoint respectively. But in the \textit{Shared} case, as the network is expected to simultaneously learn both tasks, both kinds of neurons emerge at a ratio of about $50\%$. We found that this ratio depends on the relative weight of loss terms for the two tasks. When using a different weight from the optimal in terms of maximum geometric mean accuracy, the $50\%$ ratio of specialized neuron becomes unbalanced. For a small number of in-distribution combinations, the ratio of specialized neurons may also be impacted by the relative difficulty of two tasks, with more neurons becoming specialized for the easier task (see supplement~\ref{app:SpecializationAdditionalDatasets}).

\begin{figure*}[t!]
\begin{tabular}{@{\hspace{-0.cm}}c@{\hspace{-0.1cm}}c@{\hspace{-0.1cm}}c@{\hspace{-0.1cm}}c}

\includegraphics[width=0.255\linewidth]{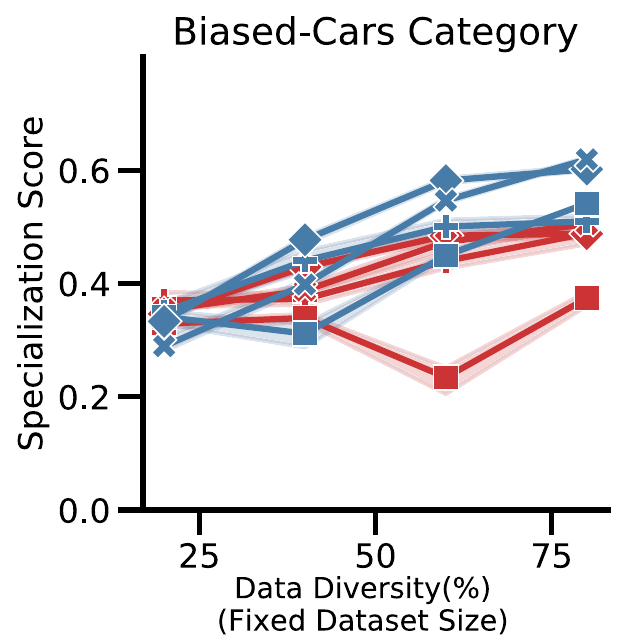}
&
\includegraphics[width=0.255\linewidth]{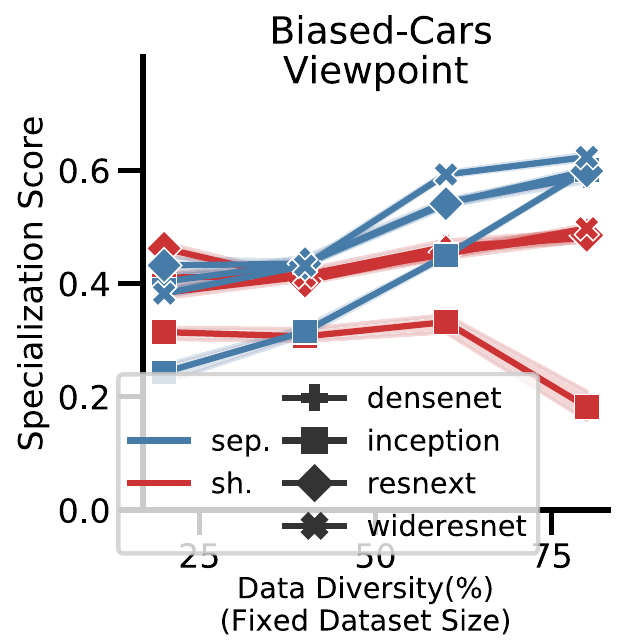}
& 
\includegraphics[width=0.255\linewidth]{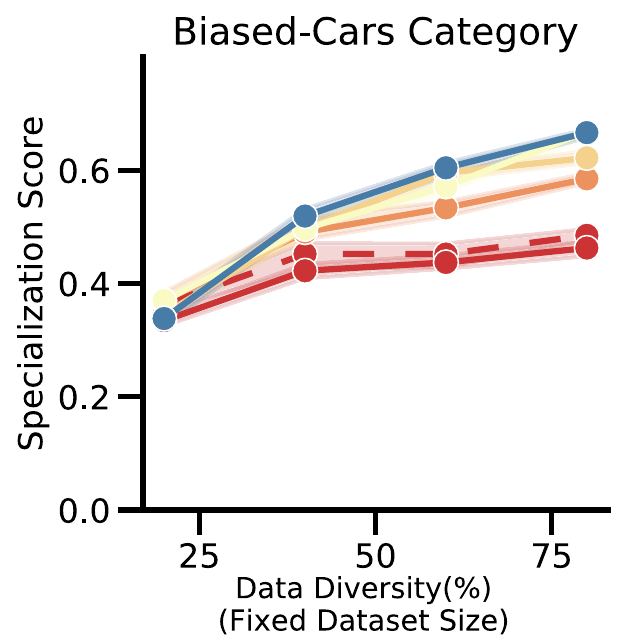}
& \includegraphics[width=0.255\linewidth]{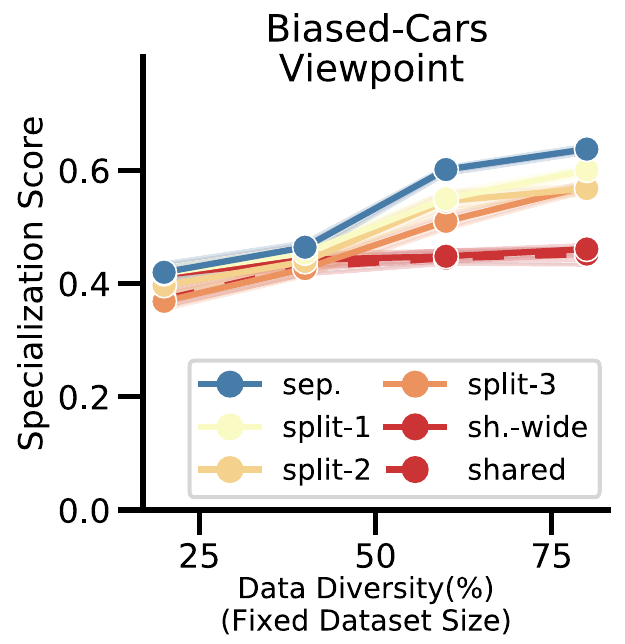}\\
(a) & (b) & (c)& (d) 
\end{tabular}
\caption{\emph{Neuron specialization (selectivity to category and invariance to viewpoint, and vice versa) in the Biased-Cars dataset.} 
(a) and (b) Median specialization score of neurons ($\Gamma^k$) in \emph{Separate} and \emph{Shared} architectures for category and viewpoint classification tasks respectively, for backbones other than ResNet-18. Confidence intervals (95\%) displayed in low opacity.
(c) and (d) Median specialization score of neurons in ResNet-18 \emph{Separate} and \emph{Shared} architectures with splits made at different blocks of the network, for category and viewpoint classification tasks respectively. }

\label{fig:specialization_scores}
\end{figure*}

In Fig.~\ref{fig:specialization_scores} we present the median of specialization scores across neurons,~\ie the median of $\Gamma^k$, in the final convolutional layer for \emph{Shared}, \emph{Split}, and \emph{Separate} architectures across multiple backbones in \textit{Biased-Cars} dataset (see supplement~\ref{app:specializationiLab} for results in other datasets). These results are presented separately for the category and viewpoint neurons. We show that as in-distribution combinations increase, there is a steady increase in the specialization score for both category and viewpoint neurons, suggesting specialization. These trends mirror the generalization trends, which suggests that specialization facilitates OOD generalization. Invariance and selectivity scores are reported separately in supplement ~\ref{app:InvarianceAndSelectivity}. We also show that specialization builds up across layers (supplement~\ref{app:SpecializationBuilds}) as expected ~\citep{goodfellow2009measuring,poggio2016visual}.

\noindent {\bf \emph{Separate} networks facilitate the emergence of specialized neurons.} Fig.~\ref{fig:specialization_scores} shows that \textit{Separate} architectures facilitate  specialization, while the \emph{Shared} architecture makes it harder for the neurons to specialize (lower specialization scores). This might be because unlike the \emph{Shared} architecture, the branches of the \textit{Separate} architecture are not forced to preserve features relevant to both tasks. Each branch can develop features which are selective to only one task, and invariant to the other. This may facilitate an increase in specialization and thus enable better performance on OOD combinations. Even though the \emph{Shared} architecture tries to split into two specialized parts, this specialization is much stronger in the \textit{Separate} architecture due to already having separate branches.

\section*{Conclusions}

We have demonstrated that CNNs generalize better to OOD category-viewpoint combinations as the training data diversity grows, for constant dataset size.  We have also shown that networks trained separately for category and viewpoint classification surpass by a large margin a shared network trained on both tasks when tested on OOD combinations. We attribute this to the branches in the \textit{Separate} architecture not being forced to preserve information about both tasks, which facilitates an increase in specialization,~\ie selectivity to category and invariance to viewpoint, and vice versa. These results are consistent across five CNN backbones and six datasets, one of them introduced in this paper as a controlled yet photo-realistic benchmark for CNN generalization. 

We also found that the aforementioned impact of data diversity and \textit{Separate} architecture are the opposite for in-distribution and OOD combinations---increased data diversity degrades in-distribution performance, and \textit{Separate} networks perform worse than \textit{Shared} ones in in-distribution combinations. This highlights that findings from  in-distribution analysis do not apply to OOD.

As a first step towards understanding generalization to OOD combinations, our work makes certain assumptions (summarized in the supplement~\ref{sec:suplLimits}) which present interesting directions for future work. These include understanding how generalization is impacted by a larger number of tasks, multiple objects in the image, object symmetries, non-rigid objects, and non-uniform ways of holding-out the test set, among others. Finally, we are intrigued to explore what other factors can help learn selective and invariant neural representations which can generalize better and lead the way towards robust, trustable CNNs.

\subsection*{Acknowledgements} 
We are grateful to Tomaso Poggio and Pawan Sinha for their insightful advice and warm
encouragement. This work has been partially supported by NSF grant IIS-1901030, a Google Faculty Research Award, the Toyota Research Institute, the Center for Brains, Minds and Machines (funded by NSF STC award
CCF-1231216), Fujitsu Laboratories Ltd. (Contract No. 40008819) and the MIT-Sensetime Alliance
on Artificial Intelligence. We also thank Kumaraditya Gupta for help with the figures, and Prafull Sharma for insightful discussions.

\subsection*{Contribution Statement} 
SM, TH, JD and XB conceived, designed and implemented the experiments and carried out the analysis, with contributions of TS, FD and HP; SM, HH, NB and FD designed and implemented the Biased-Cars dataset; SM, TS and XB wrote the manuscript with contributions of FD and HP; TS, FD, HP and XB supervised the study.

\subsection*{Conflict of Interest Statement}
This study received funding from Fujitsu Laboratories Ltd. The funder through TS had the following involvement with the study: conception of the experiment, writing of this article and supervision of the study. All authors declare no other competing interests.

\subsection*{Data and Code Availability Statement} 
Source code, demos and data are available on Github at \url{https://github.com/Spandan-Madan/generalization_biased_category_pose} (\url{https://zenodo.org/record/5636158#.YZRBKS1h1js}).

\bibliography{library}
\bibliographystyle{iclr2021_conference}

\clearpage
\appendix
\section*{Appendix}

\appendix
\renewcommand{\thefigure}{Supp.\arabic{figure}}
\renewcommand{\thetable}{Supp..\arabic{table}}
\renewcommand{\theequation}{Supp.\arabic{equation}}

\setcounter{figure}{0}
\setcounter{table}{0}
\setcounter{equation}{0}

\section{Additional details on Datasets}

\subsection{Samples from MNIST-Position and MNIST-Scale datasets}
\label{sec:supMNIST}
\begin{figure*}[!t]
\begin{tabular}{@{\hspace{0.1cm}}c@{\hspace{0.25cm}}c}
\centering\includegraphics[width=0.48\linewidth]{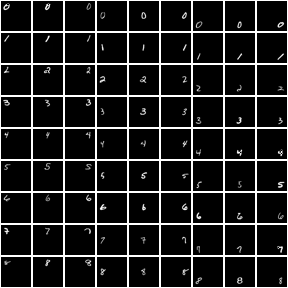}&
     \includegraphics[width=0.48\linewidth]{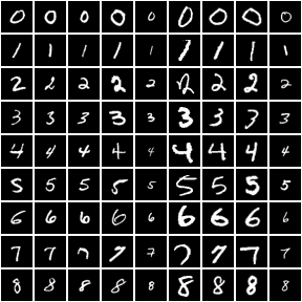}\\
     (a) MNIST-Position & (b) MNIST-Scale
\end{tabular}
 \caption{\emph{Combinations grids for MNIST-Position and MNIST-Scale.} Each row represents images from a category and each column from a viewpoint. (a) MNIST-Position was created by adding viewpoint in the form of position to images. For this, MNIST images were placed into one of nine positions in an empty three-by-three grid with equal probability. (b) MNIST-Scale was created by resizing images from MNIST to one of nine possible sizes, and then zero-padding.}
\label{fig:MNIST}
\end{figure*}

Fig.~\ref{fig:MNIST} presents one representative example for each category-viewpoint combination through the \textit{combinations grid} for the MNIST-Position and MNIST-Scale datasets.

\begin{figure*}[t!]
\begin{tabular}{@{\hspace{0.1cm}}c@{\hspace{0.1cm}}c@{\hspace{0.1cm}}c}
\centering\includegraphics[width=0.32\linewidth]{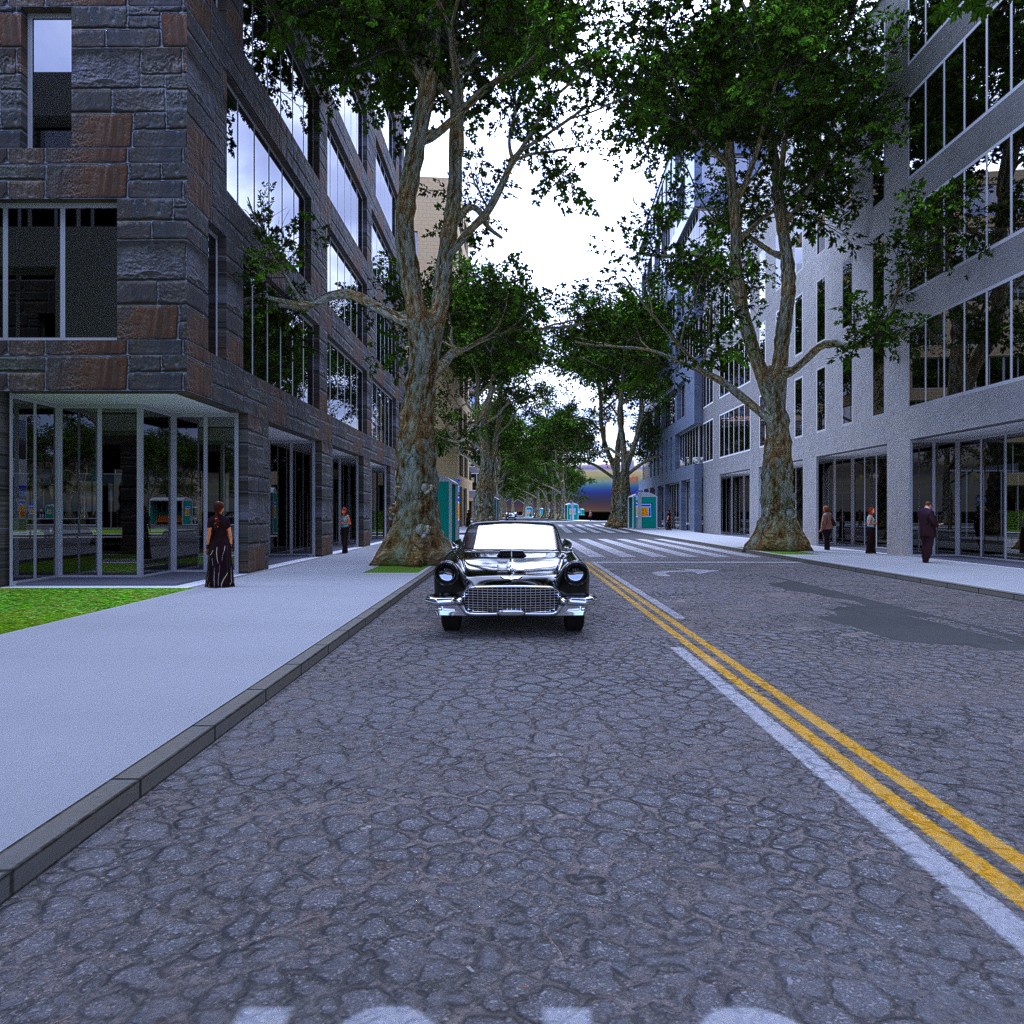}
     \includegraphics[width=0.32\linewidth]{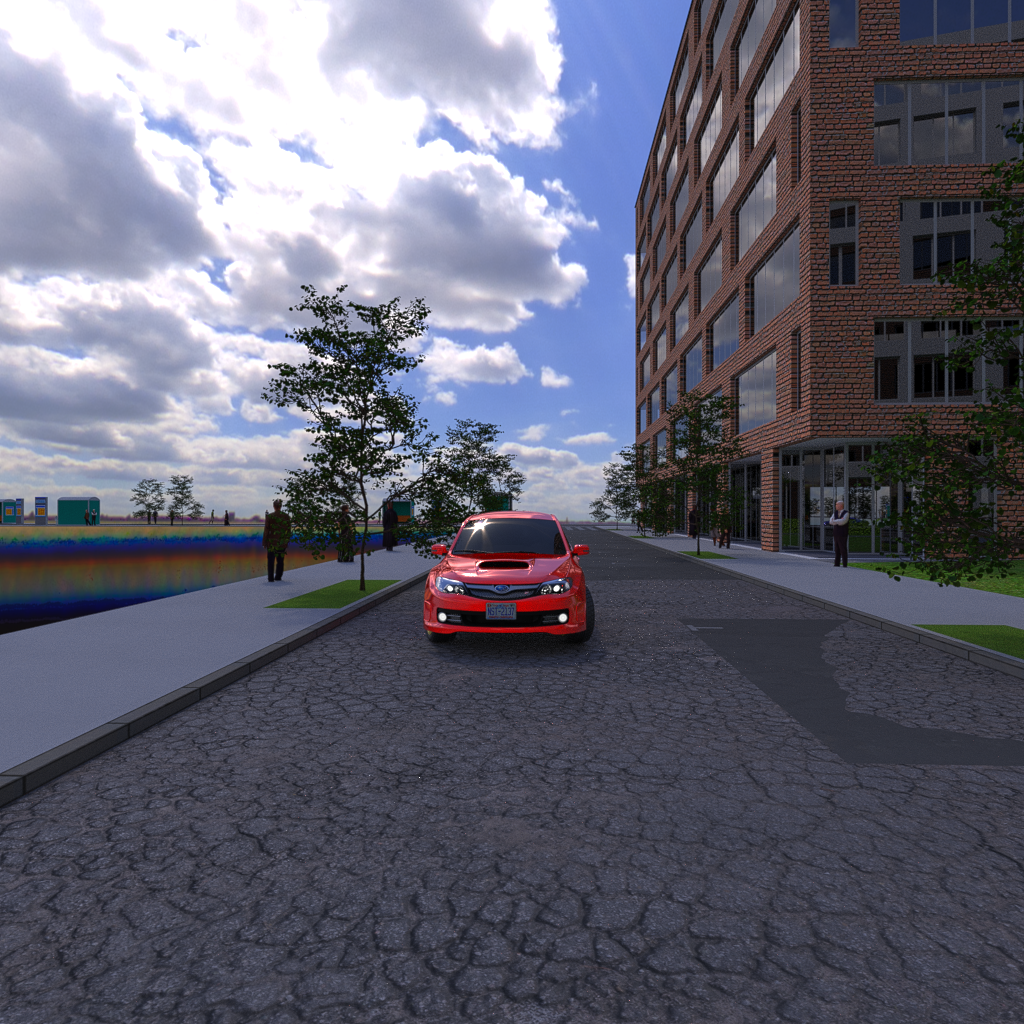}
& 
\includegraphics[width=0.32\linewidth]{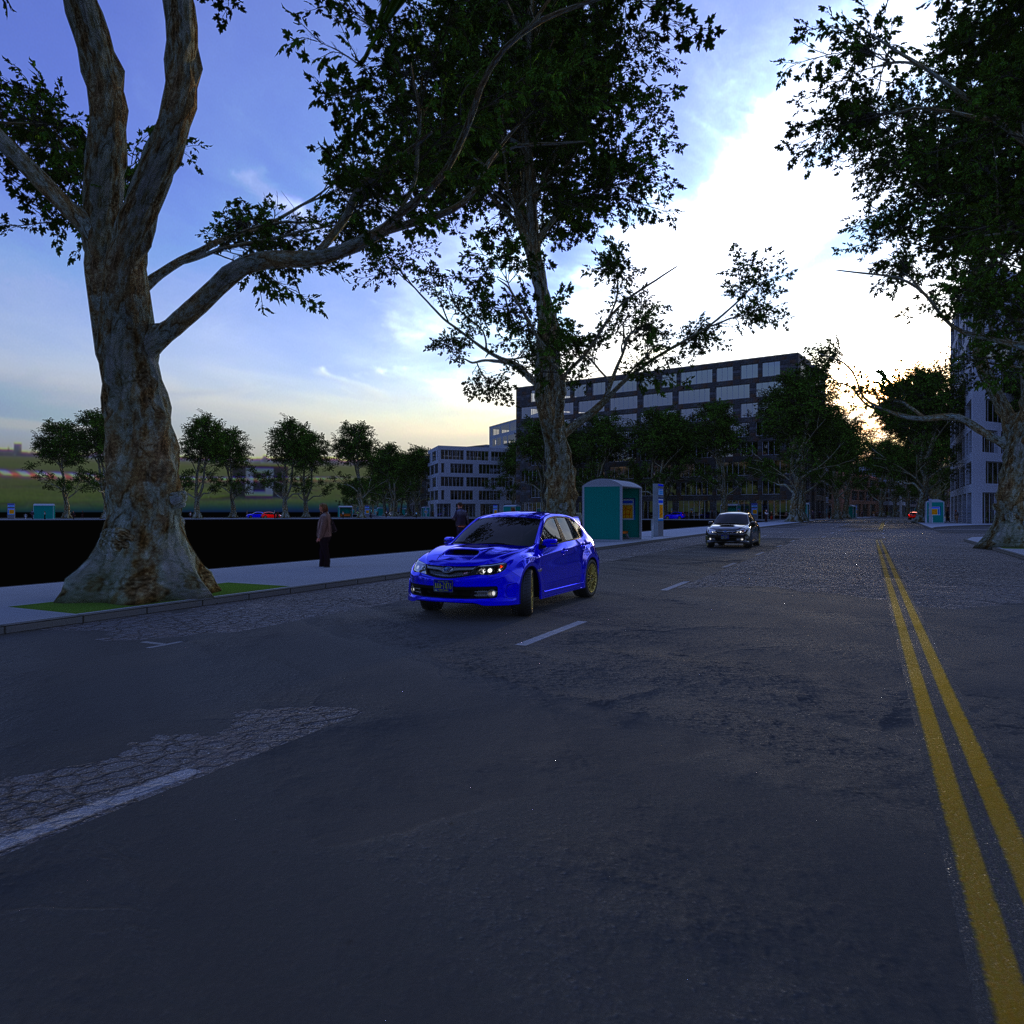}\\
\includegraphics[width=0.32\linewidth]{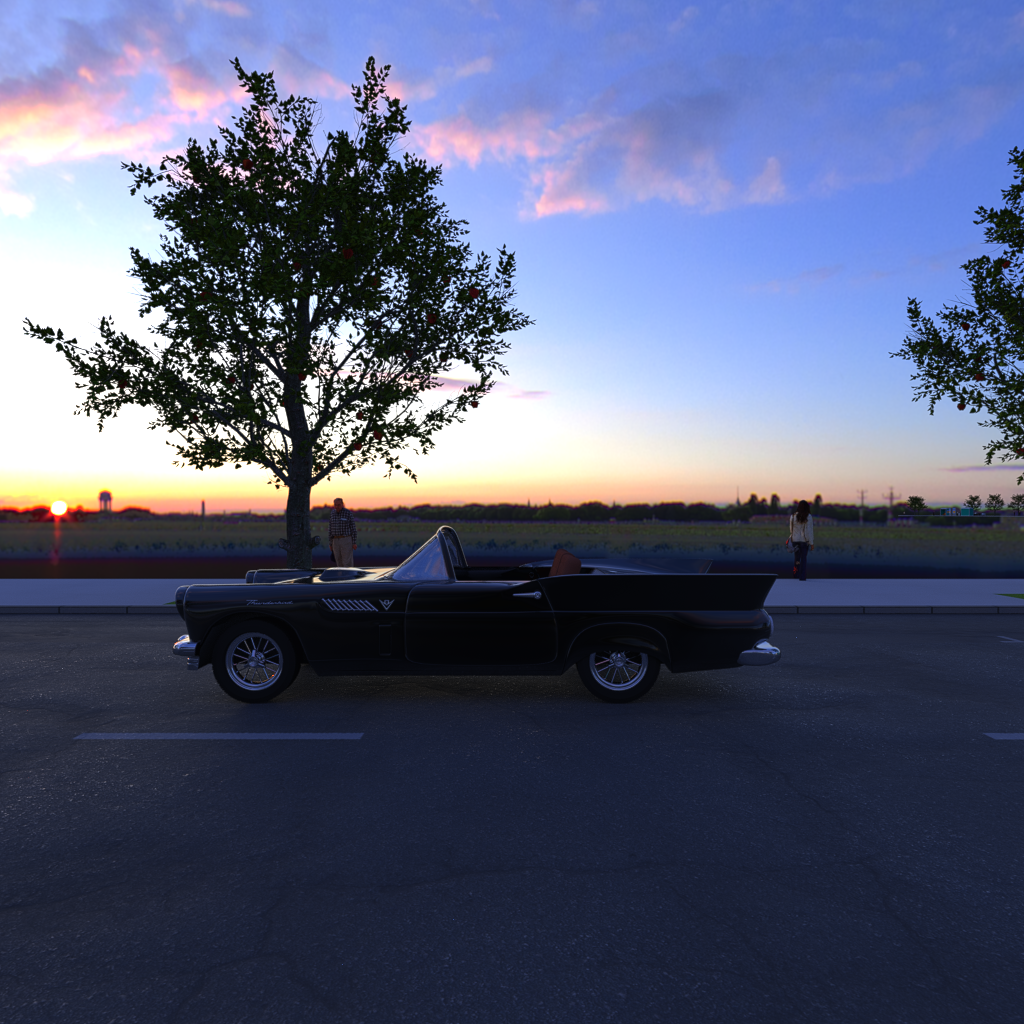}
\includegraphics[width=0.32\linewidth]{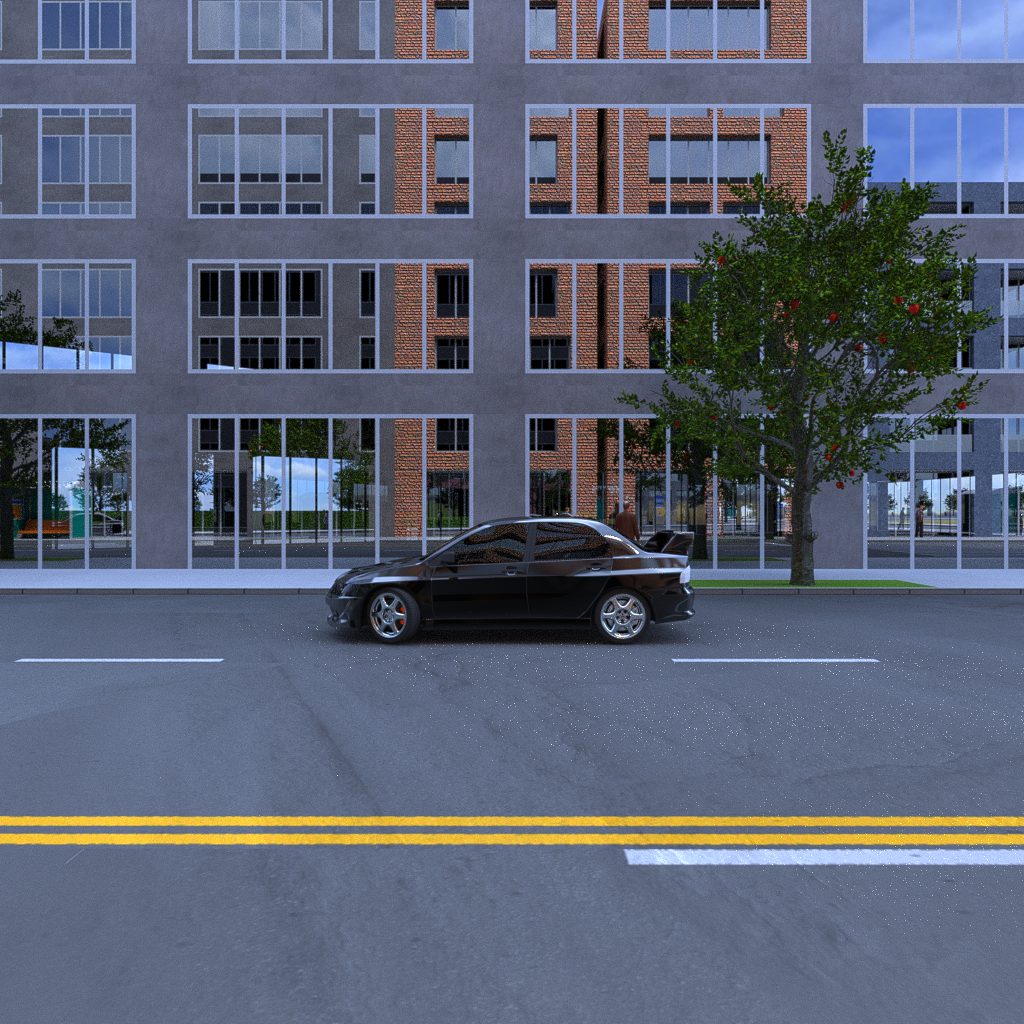}
& \includegraphics[width=0.32\linewidth]{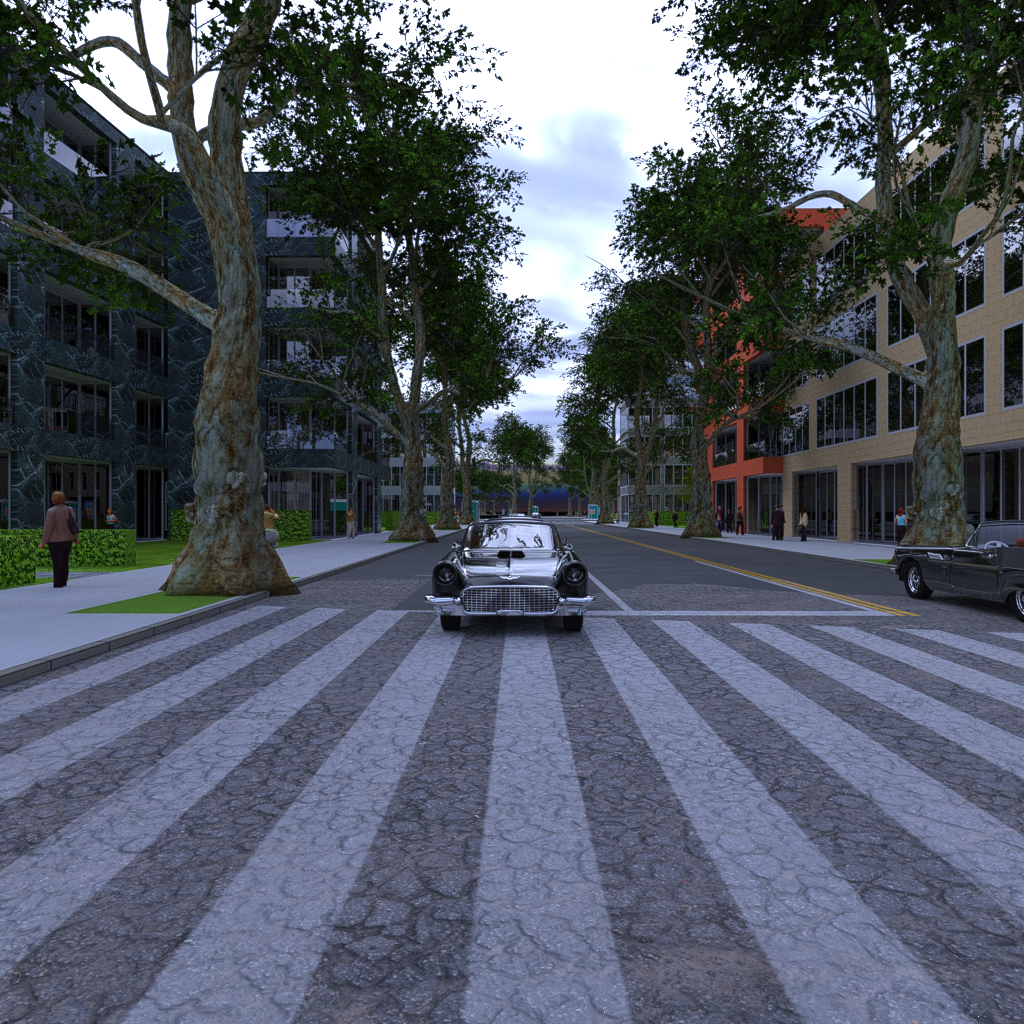}\\
\includegraphics[width=0.32\linewidth]{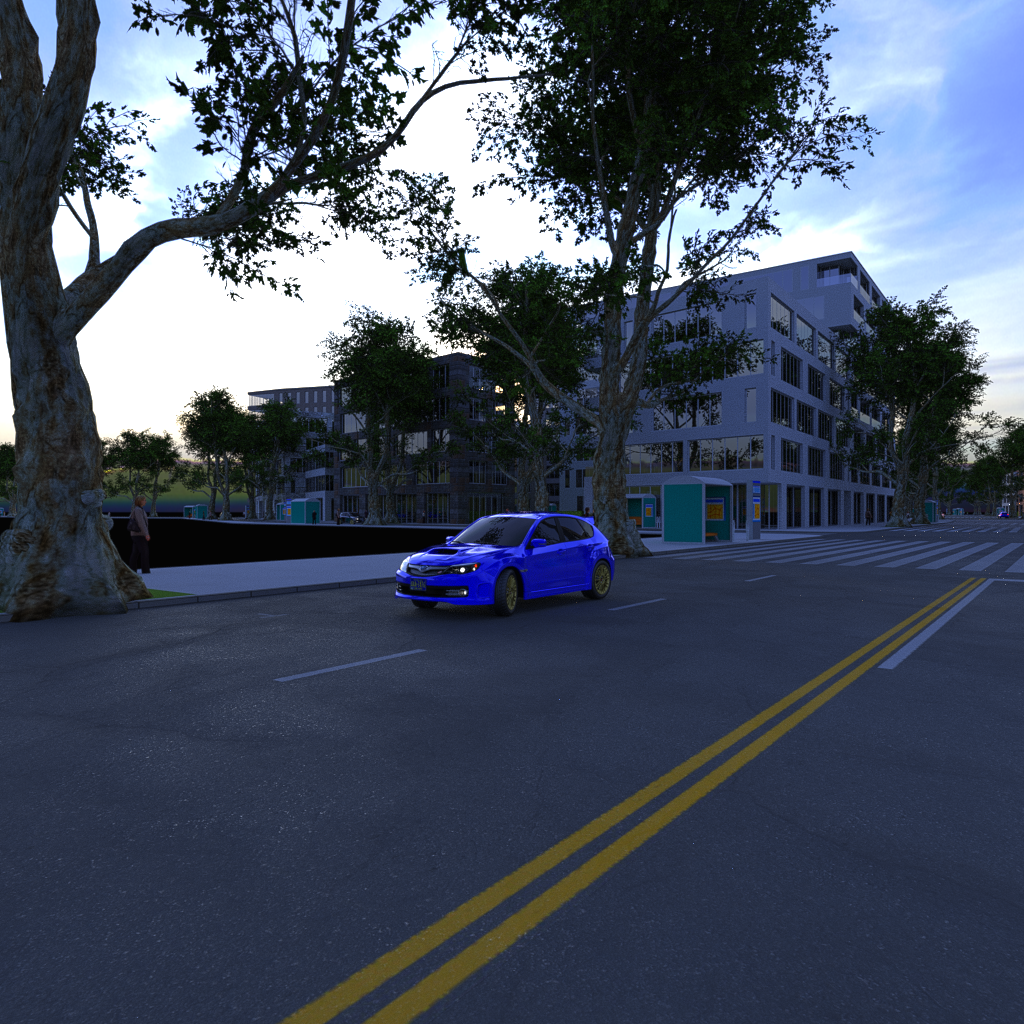}
     \includegraphics[width=0.32\linewidth]{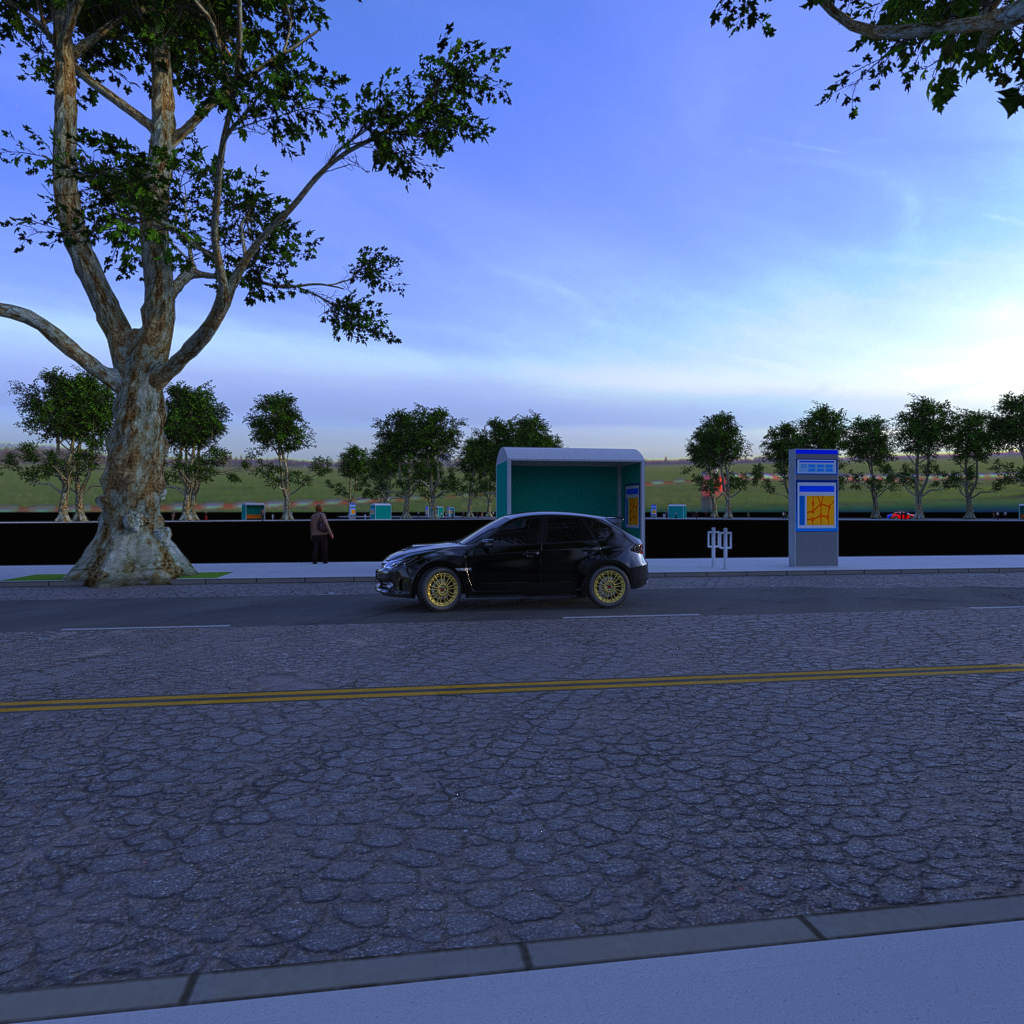}
& \includegraphics[width=0.32\linewidth]{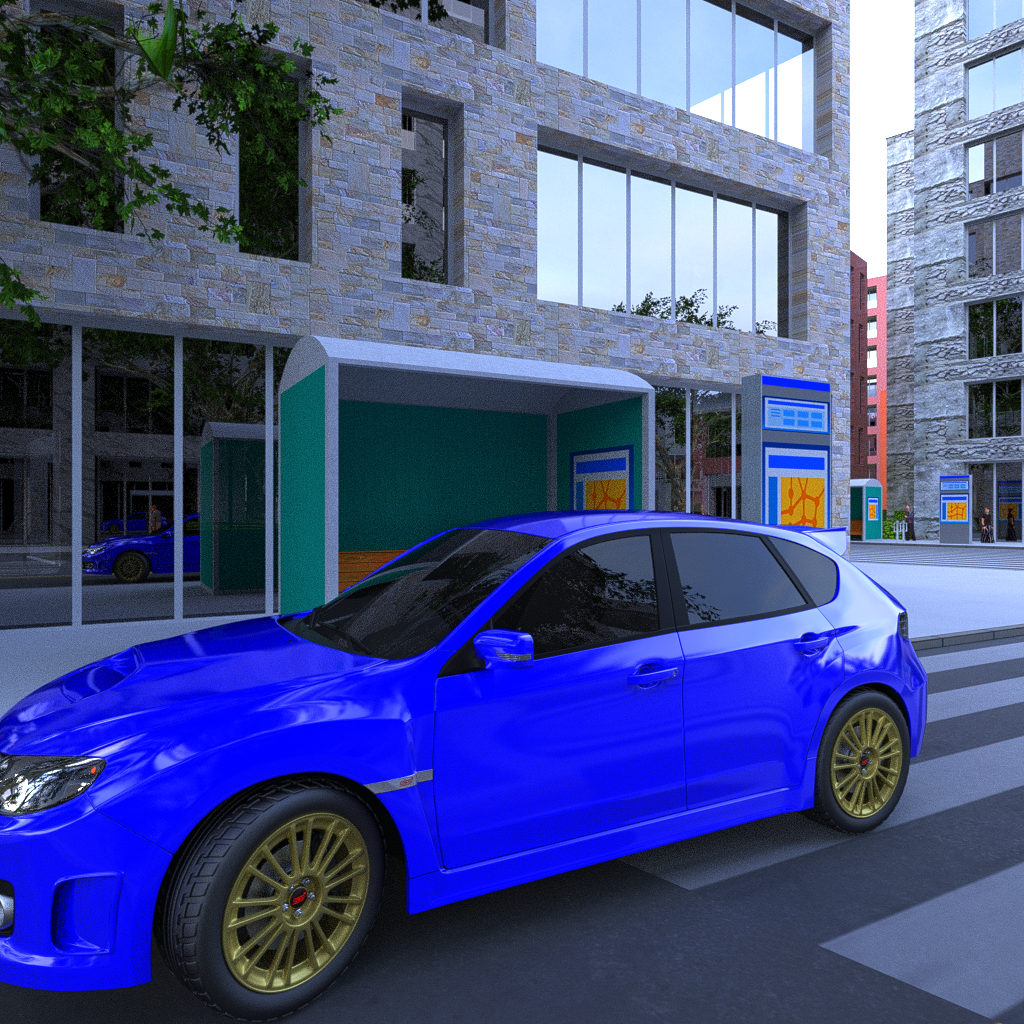}\\
\includegraphics[width=0.32\linewidth]{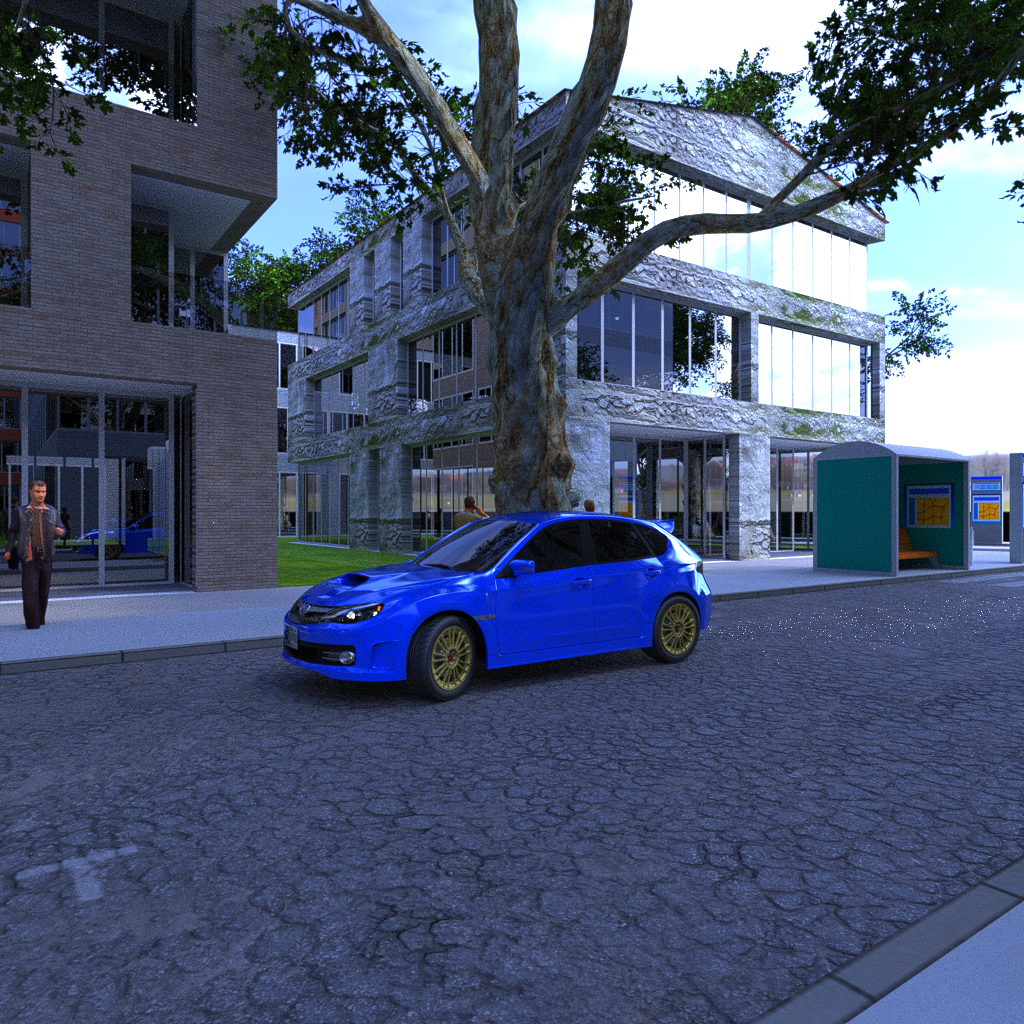}
\includegraphics[width=0.32\linewidth]{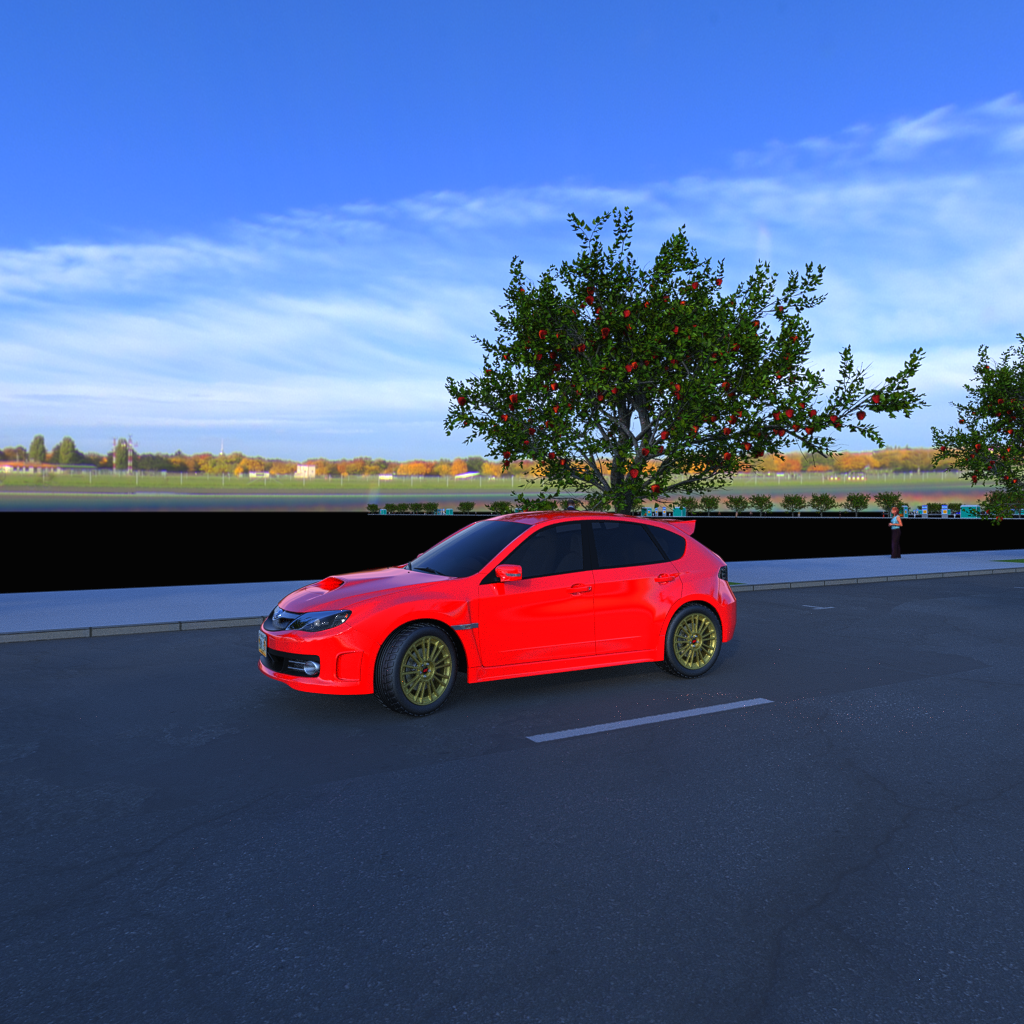}
&\includegraphics[width=0.32\linewidth]{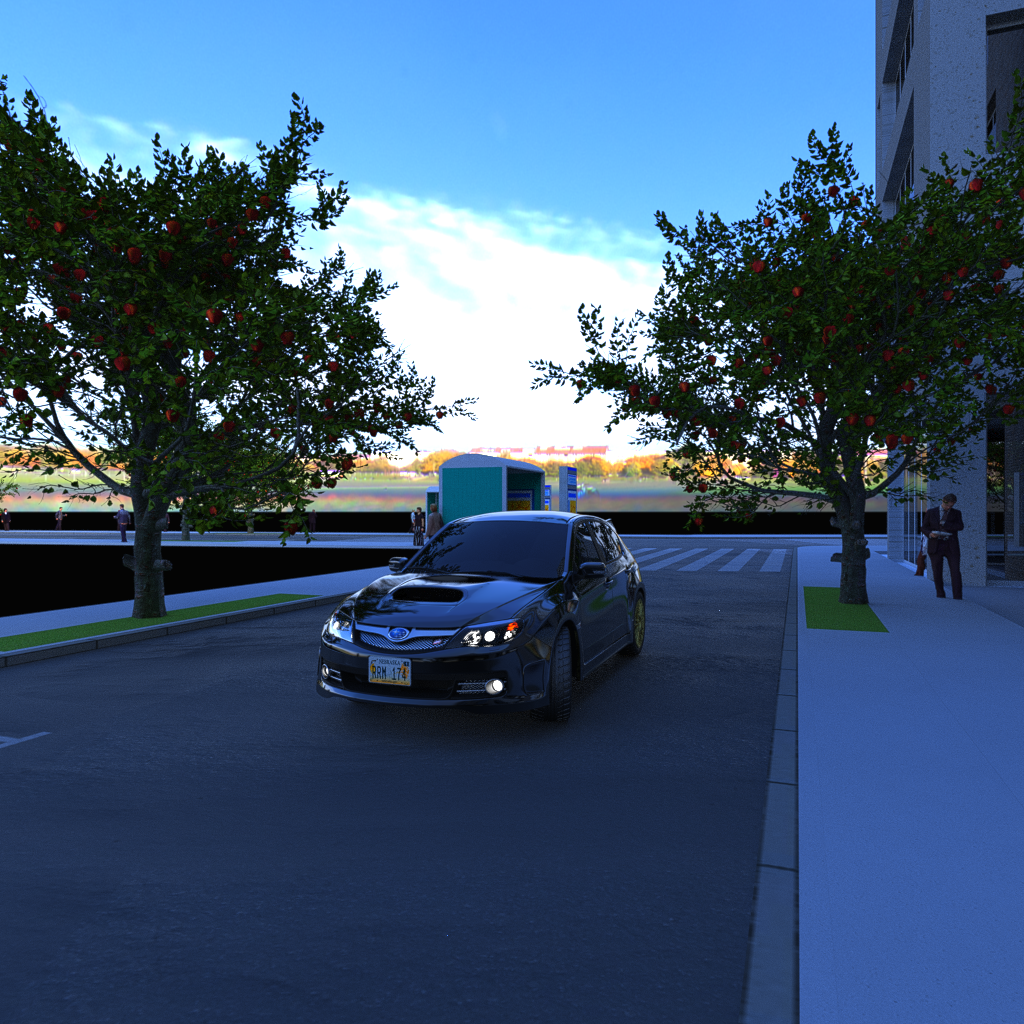}\\
\includegraphics[width=0.32\linewidth]{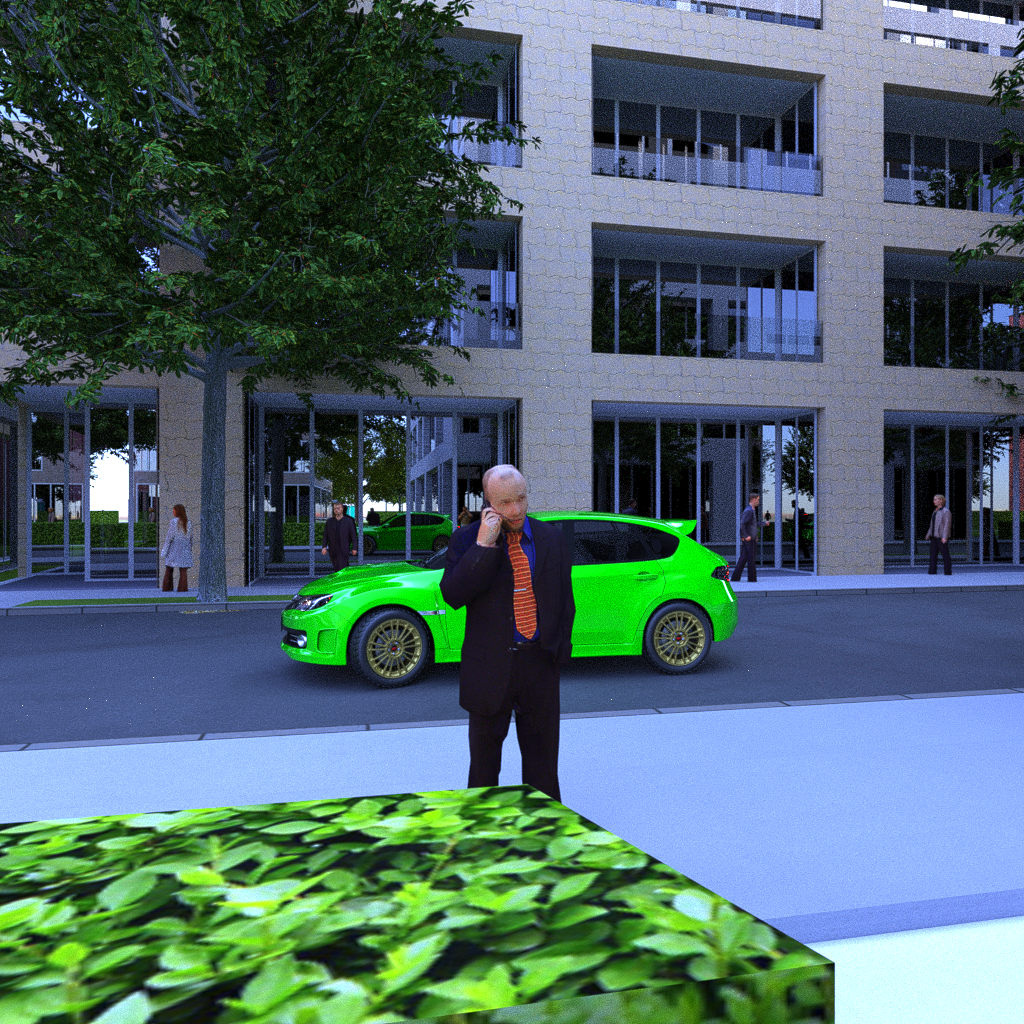}
     \includegraphics[width=0.32\linewidth]{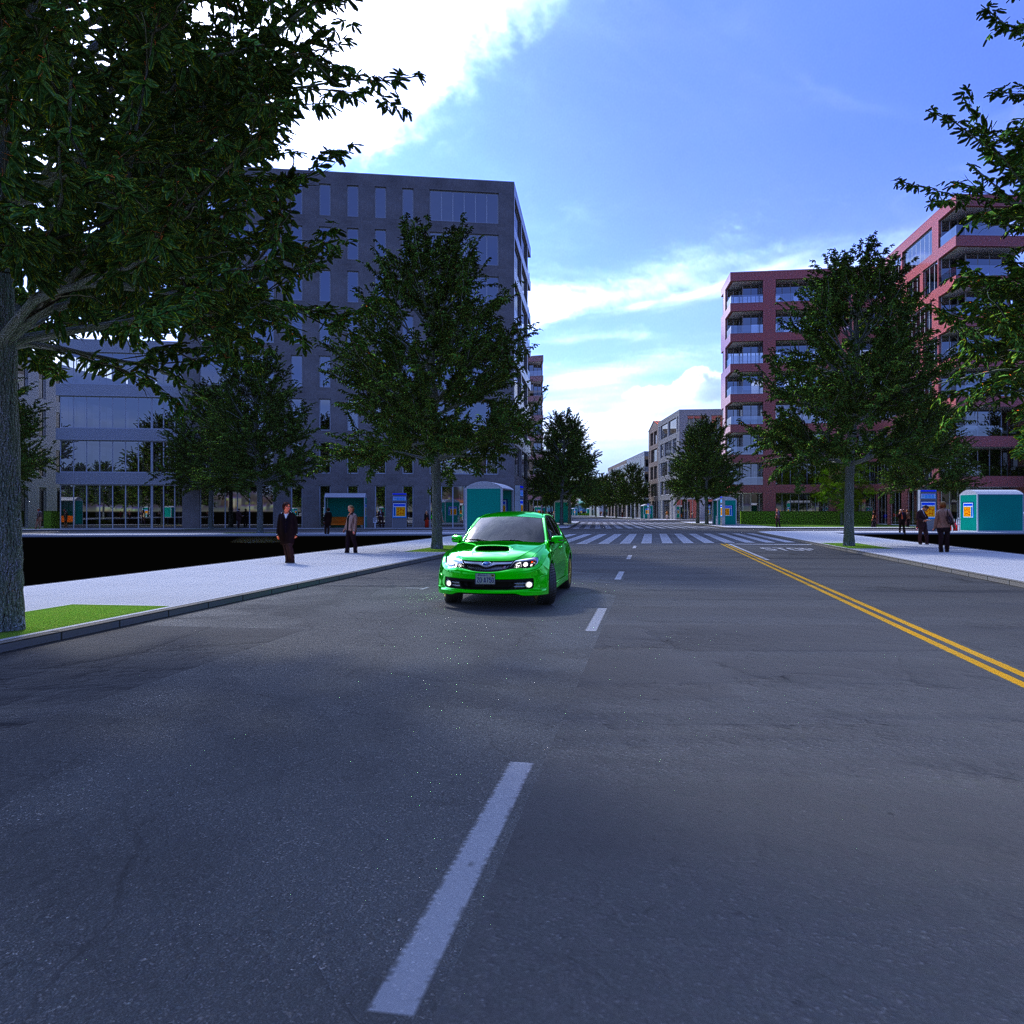}
& \includegraphics[width=0.32\linewidth]{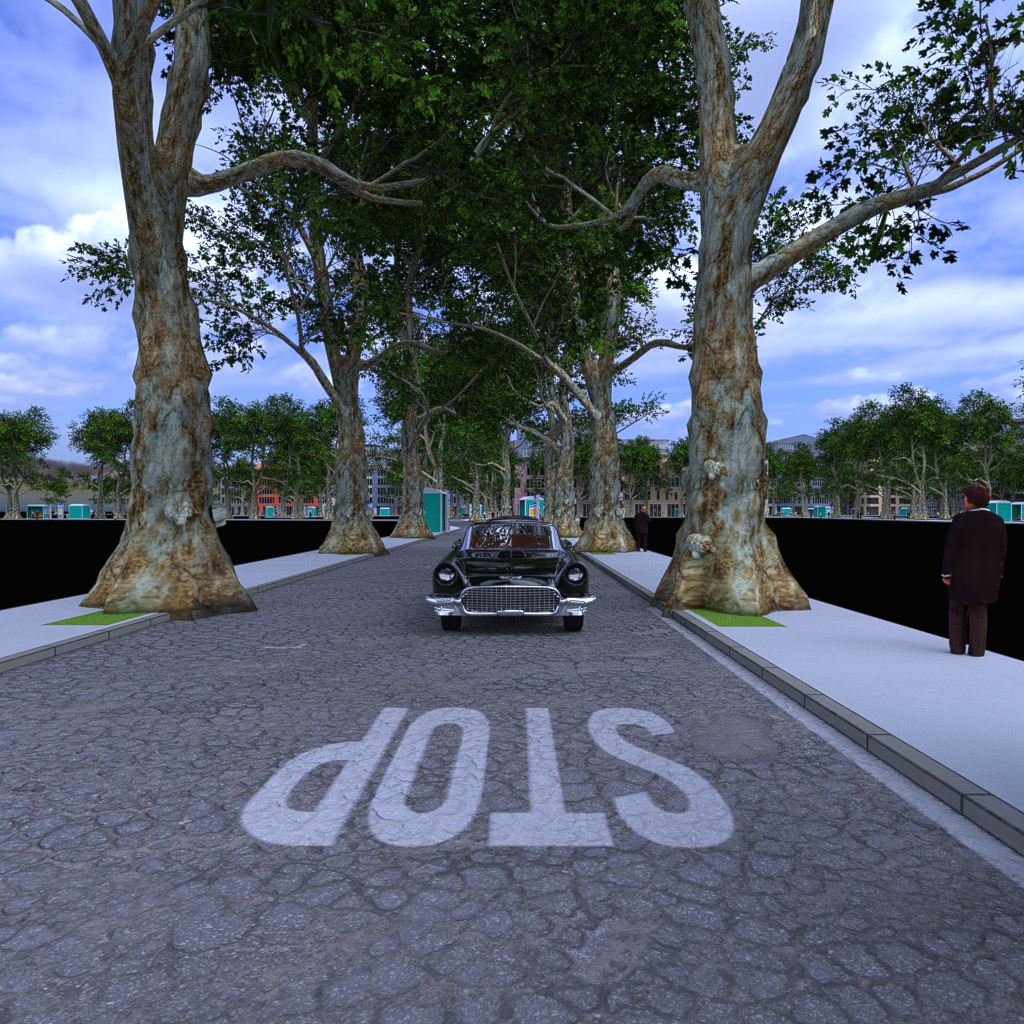}
\end{tabular}
\caption{\emph{Sample images from the Biased-Cars dataset.}}
\label{fig:samples_biased_cars}
\vspace{-0.3cm}
\end{figure*}

\subsection{Rendering Pipeline for \textit{Biased-Cars} Dataset}
\label{sec:suppCars}

To generate photo-realistic data with systematic, controlled biases we implemented our computer graphics pipeline which offered us fine grained control over scene attributes including but not limited to - backgrounds, textures, lighting and geometry. Below we present the details of our rendering pipeline, along with some sample images.

\noindent{\textbf{Pipeline Details:}} We used Esri CityEngine~\citep{EsriCityEngine} to model the city layout and geometry, to which we add 3D assets - car models, pedestrians, trees, street furniture like bus stops, textures for buildings, roads and car paints. Blender Python API~\citep{blender} is used to modify the 3D city file. This includes placing vehicles and other assets at user defined locations, modifying their material properties including vehicle paint, adding specified textures to roads, buildings and pedestrians, and defining camera attributes (lens, field of view, motion blur etc) and camera locations. For randomization, a distribution over each parameters was defined. For instance, a discrete uniform distribution over possible car color paints. Similarly, we defined distributions over object positions in the city, camera viewpoint and distance, among other factors.

Sample images are shown in Fig.~\ref{fig:samples_biased_cars} below, rendered at $1024 \times 1024$ pixels. As network input was $224 \times 224$, training images were rendered at $256 \times 256$ and then resized to $224 \times 224$ (as side length of the form $2^k$ lead to computational gains in physically based rendering). Physically based rendering accurately models the flow of light in the scene resulting in highly photo-realistic images. As can be seen, our pipeline reproduces lighting artefacts like color bleeding and specular highlights very gracefully. As shown, images include cars seen from different distances and viewpoints, under different lighting conditions, scene clutter and even occlusions.

\FloatBarrier

\section{Selectivity and Invariance}
In the paper we defined the selectivity score of a neuron with respect to category and its invariance score with respect to viewpoint. Following the same notation as the paper: $a_{ij}^k$ denotes the \textit{activations grid} for neuron $k$, where each row represents one category and each column represents a viewpoint.

\subsection{Normalization of \textit{activations grid}}
\label{sec:supl:Norm}

For every neuron, we first normalize its activations for every image by dividing them by its maximum activation across all images. This ensures that that the activation for every image lies between $0$ and $1$ for all neurons. The entries of the \textit{activations grid} for a neuron are then computed by averaging these normalized activation for images belonging to each category-viewpoint combination.

The \textit{activations grid} is then normalized to be between $0$ and $1$. To do so, we subtract the minimum of the \textit{activations grid} and then divide it by the maximum.

\subsection{Selectivity and Invariance with respect to viewpoint}
In the paper, we used $i^{\star k}$, $S^k_c$, $I^k_v$ to denote the \textit{preferred category}, selectivity score with respect to category and invariance score with respect to viewpoint respectively. We also presented these equations to compute these quantities:

\begin{equation}
i^{\star k} = {\arg \max_i} \sum_{j} a_{ij}^k.
\end{equation}

\begin{equation}
     S^k_c = \frac{\hat{a}^k - \bar{a}^k}{\hat{a}^k + \bar{a}^k},
     \; \;  \; \; \mbox{where}  \;  \hat{a}^k = \frac{1}{N}\sum_{j} a_{i^{\star k} j}^k ,\;\; \bar{a}^k = \frac{\sum_{i\neq i^{\star k}}\sum_j {a}_{ij}^k}{N(N-1)}.\\
\end{equation}

\begin{equation}
     I^k_v = 1 -\Big(\underset{j}{\max} \;\; a_{i^{\star k}j}^k - \underset{j}{\min} \;\; a_{i^{\star k}j}^k\Big)
 \end{equation}

We now present how to compute the selectivity with respect to viewpoint, and invariance with respect to category, denoted as $S^k_v$ and $I^k_c$ respectively. These can be obtained by first finding the \textit{preferred viewpoint}, denoted as $j^{\star k}$, and proceeding as in the above equations:

\begin{equation}
j^{\star k} = {\arg \max_j} \sum_{i} a_{ij}^k.
\end{equation}

\begin{equation}
     S^k_v = \frac{\hat{a}^k - \bar{a}^k}{\hat{a}^k + \bar{a}^k},
     \; \;  \; \; \mbox{where}  \;  \hat{a}^k = \frac{1}{N}\sum_{i} a_{ij^{\star k}}^k ,\;\; \bar{a}^k = \frac{\sum_{j\neq j^{\star k}}\sum_i {a}_{ij}^k}{N(N-1)}.\\
\end{equation}

\begin{equation}
     I^k_c = 1 -\Big(\underset{i}{\max} \;\; a_{ij^{\star k}}^k - \underset{i}{\min} \;\; a_{ij^{\star k}}^k\Big)
 \end{equation}
 
Observe that like $S^k_c$, $S^k_v$ is a value between $0$ and $1$, and higher value indicates that the neuron is more active for the \emph{preferred viewpoint} as compared to the rest of the viewpoints. $I^k_c$ too is a value between $0$ and $1$, with higher values indicating higher invariance to the category for images containing the \textit{preferred viewpoint}.

\clearpage
\section{Experimental Details and Hyper-Parameters}
\label{sec:suplDetails}

Each of our four datasets contains both category and viewpoint labels for all images. We define the location and the scale as the viewpoint for MNIST-Position and MNIST-Scale datasets respectively. For both iLab and \textit{Biased-Cars} dataset, the viewpoint refers to the azimuth viewpoint. Networks are trained to classify both category and viewpoint labels simultaneously, and all models are trained from scratch, without any pre-training to ensure controlled testing. This ensures that any existing biases in common pre-training datasets like ImageNet~\citep{torralba2011unbiased} do not impact our results.

\noindent{\textbf{Number of Images:}} The number of training images is kept fixed for every dataset, and was decided by training networks on these datasets while gradually increasing size, till the performance on OOD combinations saturated. For the \textit{Biased-Cars} dataset, performance plateaud at 3,400 train, 445 validation, and 800 OOD test images. For iLab, we used 70,000 train, 8,000 validation images, and 8,000 OOD test images. As the iLab dataset is a natural image dataset, it required much more images to saturate. For MNIST, 54,000 train, 8,000 validation and 8,000 test images were used. 

\noindent{\textbf{Hyper-parameters}}: We used the Adam~\citep{kingma2014adam} optimizer with 0.001 as learning rate, and ReLU activations. For the \textit{Biased-Cars} datasets, all models were trained for 200 epochs, while we trained for 50 epochs for the iLab dataset. MNIST-Position and MNIST-Scale were trained for 5 epochs. These stopping criterion were picked to ensure convergence on generalization to OOD combinations. All experiments were repeated multiple times and confidence intervals (95\%) are shown in the plots in the main paper. iLab and \textit{Biased-Cars} experiments were repeated 3 times each, and MNIST experiments were repeated 10 times. Loss for training \textit{Shared} architectures was simply the sum of CrossEntropy Loss for both category and viewpoint classification. We compared how different weighted sums perform, and found this to be performing best as measured by the geometric mean of category and viewpoint classification.

\newpage
\section{Additional Experiments:``When Do CNNs generalize to OOD combinations?''}
\label{sec:suplExperiments}
Below we present additional results that re-inforce our findings presented in the results sections of the main paper. 

\FloatBarrier
\begin{figure*}[!t]
\centering\includegraphics[width=0.5\linewidth]{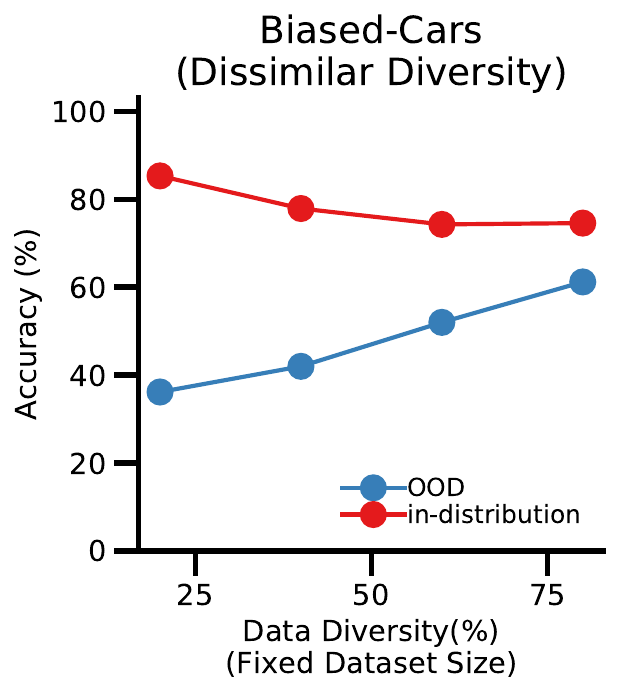}

 \caption{\emph{Generalization of \emph{Separate} architectures to OOD combinations as number of in-distribution combinations are increased while ensuring that viewpoints increasingly distant from the OOD set are added (Dissimilar Diversity).} To control that the increasing generalization performance as the data diversity increases is not due to closer viewpoint angles between In-distribution and OOD combinations, we created a dataset split where the In-distribution combinations start closest to the OOD combinations (in terms of viewpoint angle), and become increasingly distant. Thus, as In-distribution combinations are increased, the training data becomes increasingly dissimilar to the OOD combinations (test set). The results show that the performance on the OOD combinations still improves as data diversity increases. This experiment discards the hypothesis the increase in generalization performance is due to having closer viewpoint angles between the In-distribution and OOD combinations}.
\label{fig:baised_cars_extrapolation_set}
\end{figure*}

\subsection{Similarity between In-Distribution and OOD Combinations}

\label{sec:suplSimilarity}

To discard that the increasing generalization performance as the data diversity is increased is not due to having closer viewpoint angles between the in-distribution and OOD combinations, we provide the following control experiment.
For the smallest number of in-distribution combinations, we use the combinations that are the closest to the OOD combinations (\ie consecutive bins). As we increase the number of in-distribution combinations, we keep adding the rest of in-distribution combinations in the order of closeness to the OOD combinations. Fig.~\ref{fig:baised_cars_extrapolation_set} show a clear increase of the accuracy in OOD combinations. The increase of accuracy in this experiment can not be explained by the fact that the in-distribution combinations tend to be more similar to the OOD combinations when increasing the data diversity, because in this experiment, the combinations tend to be more dissimilar as increasing the data diversity. Thus, this experiment discards that the increase in generalization performance is due to having closer viewpoint angles between the in-distribution and OOD combinations

\FloatBarrier

\subsection{Separate performance of category and viewpoint classification}

\label{sec:suplSeparateAcc}

\begin{figure*}[t!]
\begin{tabular}{@{\hspace{-0.cm}}c@{\hspace{-0.1cm}}c@{\hspace{-0.1cm}}c@{\hspace{-0.1cm}}c}
     \includegraphics[width=0.255\linewidth]{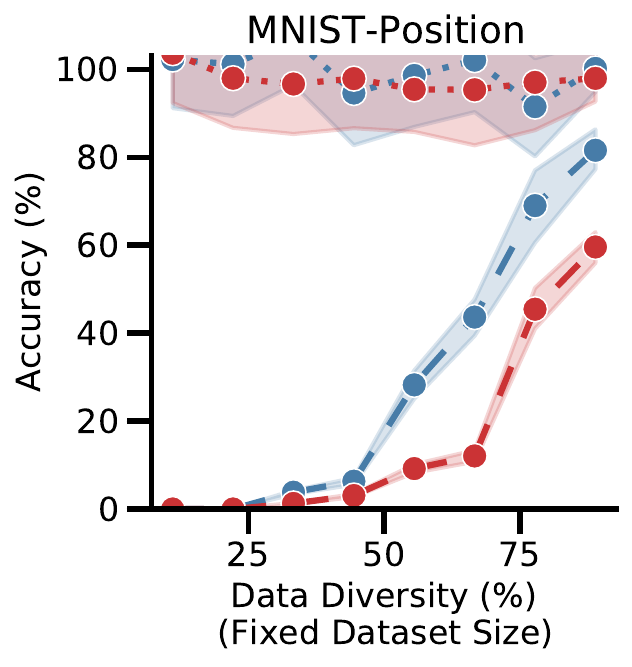}
&
     \includegraphics[width=0.255\linewidth]{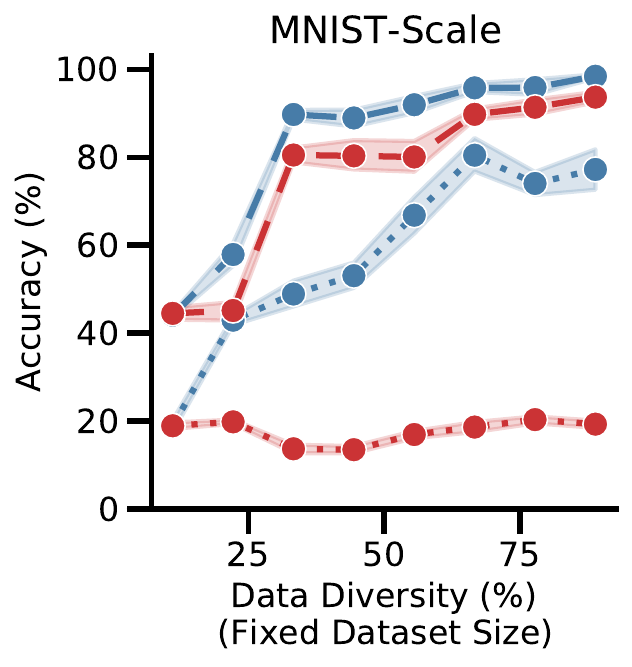}
&  \includegraphics[width=0.255\linewidth]{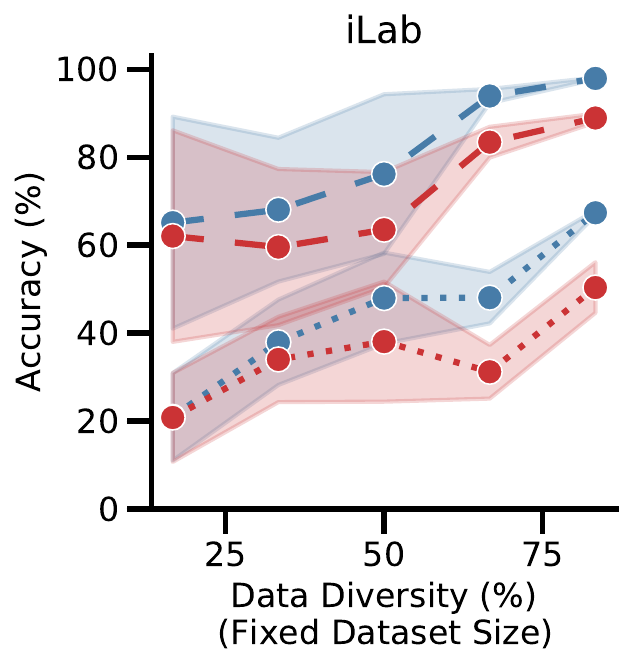} &  \includegraphics[width=0.255\linewidth]{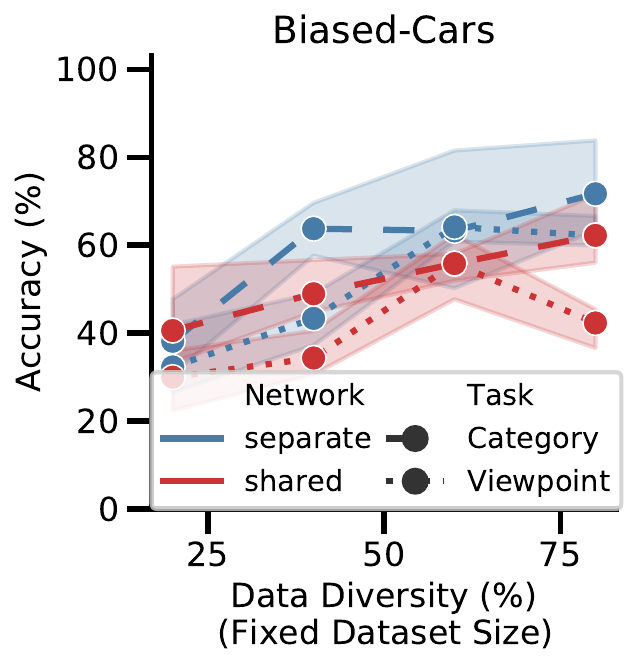}  \\
(a) & (b) & (c)& (d) \\
\end{tabular}
\caption{\emph{Generalization performance for category recognition and viewpoint estimation for \textit{Shared} and \textit{Separate} ResNet-18 as in-distribution combinations are increased for all datasets.} The category recognition accuracy and viewpoint estimation accuracy are reported along with confidence intervals (95\%) (a)~MNIST-Position dataset. (b)~MNIST-Scale dataset. (c)~iLab dataset. (d)~Biased-Cars dataset.}
\label{fig:accuracy_tasks}
\end{figure*}

In Fig.~\ref{fig:accuracy_tasks}, we show the individual accuracy for category and viewpoint classification in OOD category-viewpoint combinations. The results show that \emph{Separate} also obtains better accuracy than \emph{Shared} for each individual task accuracy. Note that the relative accuracy of the two tasks varies depending on the dataset, and no task is consistently harder than the other across all datasets. For instance, viewpoint classification is easier for MNIST-Position, while it is significantly hard for MNIST-Scale. MNIST digits are centered by default, and when placed in different positions to create MNIST-Position images, the viewpoint is easily distinguishable. For MNIST-Scale however, there is little visual variation between adjacent scales, which leads to a poor Top-1 classification accuracy for viewpoint (scale) classification.


Furthermore, we have found that for MNIST-Position, the pooling operation at the end of ResNet-18 is critical to obtain good generalization accuracy to OOD category-viewpoint combinations.  We evaluated ResNet-18 without the pooling operation and the category recognition accuracy of OOD category-viewpoint combinations dropped to baseline. Pooling facilitates an increase of position invariance and it does not harm the viewpoint classification accuracy (as shown by \cite{azulay2019deep}, pooling does not remove the position information).

\FloatBarrier

\subsection{Number of neurons in shared \emph{vs.} separate networks}

\label{app:NumberNeurons}
\begin{figure*}[!t]
\begin{tabular}{@{\hspace{0.1cm}}c@{\hspace{0.1cm}}c}
\centering\includegraphics[width=0.5\linewidth]{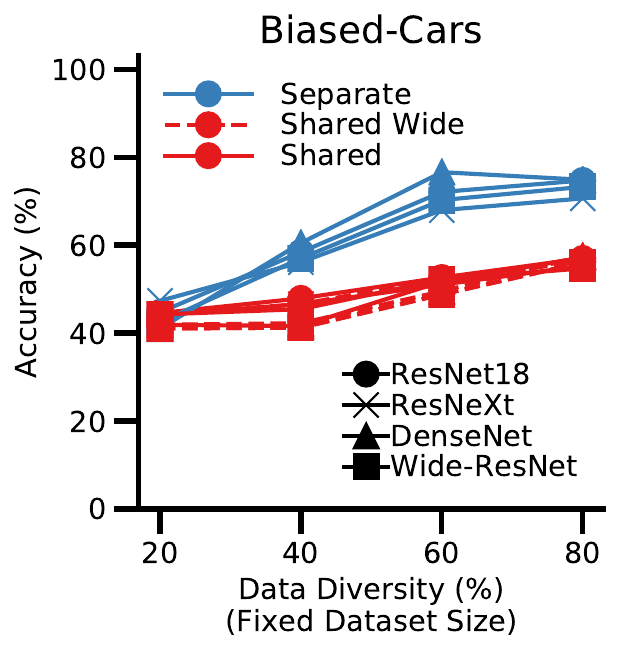}&
     \includegraphics[width=0.5\linewidth]{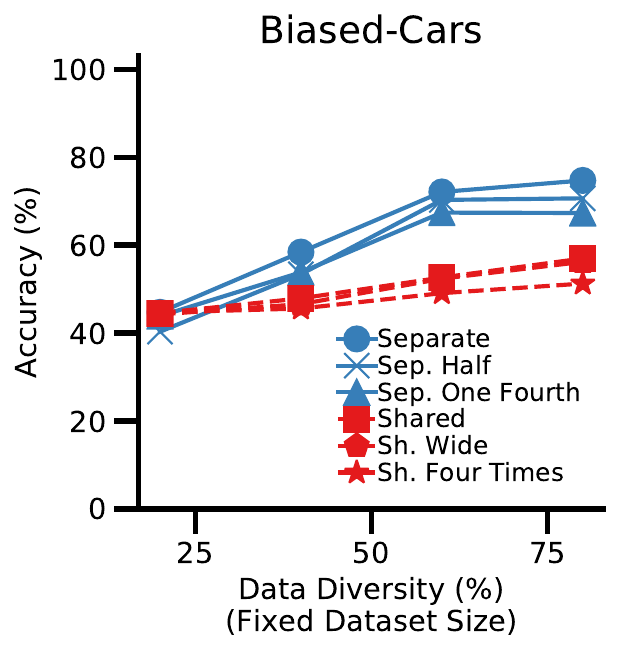}
\end{tabular}
 \caption{\emph{Generalization to OOD combinations as number of neurons per layer are varied for the ResNet-18 backbone.} \textit{Separate} architectures substantially outperform \textit{Shared} architectures across a range of widths, \ie number of neurons per layer. The \textit{Separate} architecture contains double the parameters as the \textit{Shared} architecture, as there is no weight sharing in the \textit{Separate} case. Variants of these architectures are created by increasing or decreasing the neurons in each layer by a factor of 2 at a time. Even at one-eighth the number of neurons, the \textit{Separate One Fourth} architecture generalizes much better to OOD combinations as compared to the \textit{Shared Four Times} architecture.}
\label{fig:width_control}
\end{figure*}

To control for the number of neurons in \textit{Shared} and \textit{Separate} architectures, we present additional results with the \textit{Biased-Cars} dataset in Fig.~\ref{fig:width_control}. In the paper, we presented the \textit{Shared-Wide} architecture for the ResNet-18 backbone, which is the \textit{Shared} architecture with double the number of neurons per layer, \ie double the width. Here we go one step further and test a number of similar scenarios with the ResNet-18 backbone. The \textit{Separate Half} and \textit{Separete One Fourth} architectures are made by reducing the number of neurons in every layer to one half, and one fourth of the original number respectively. It is to be noted, that the \textit{Separate} architectures has double the number of neurons as the \textit{Shared} architecture, as there is no weight sharing between branches in the \textit{Separate} case. Thus, the \textit{Separate Half} architecture has the same number of neurons as the \textit{Shared} architecture, and the \textit{Separate} architecture has the same number as the \textit{Shared-Wide} architecture. In a similar vein, the \textit{Shared Four Times} was created by multiplying the neurons in each layer of the \textit{Shared} architecture four times. Thus, the \textit{Shared Four Times} has double the number of neurons as compared to the \textit{Shared Wide} architecture, and 4 times the \textit{Shared} architecture.

As can be seen in Fig.~\ref{fig:width_control}, even at one-eighth number of neurons, the \textit{Separate One Fourth} architecture substantially outperforms the \textit{Shared Four Times} architecture at generalizing to OOD category-viewpoint combinations. This confirms that our findings are not a function of the number of neurons in the \textit{Shared} and \textit{Separate} architectures.

\FloatBarrier

\subsection{Number of Training examples}
\label{app:NumberTraining}
\begin{figure*}[!t]
\begin{tabular}{@{\hspace{0.1cm}}c@{\hspace{0.1cm}}c}
\centering\includegraphics[width=0.5\linewidth]{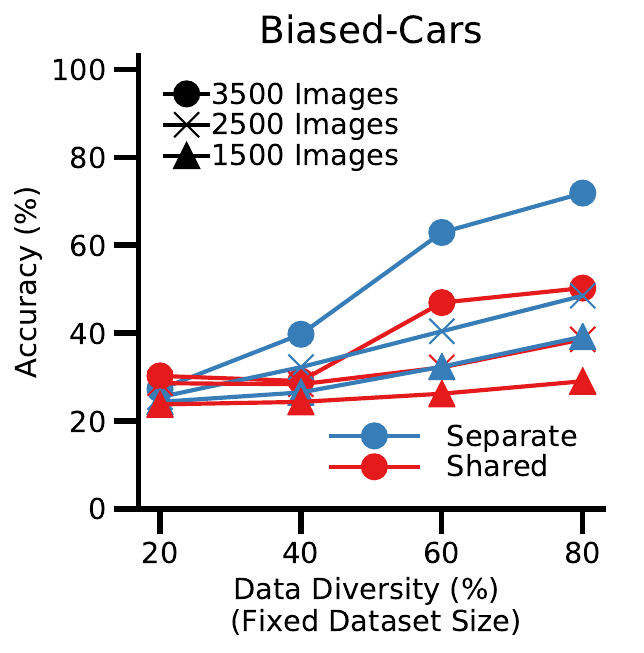}&
     \includegraphics[width=0.5\linewidth]{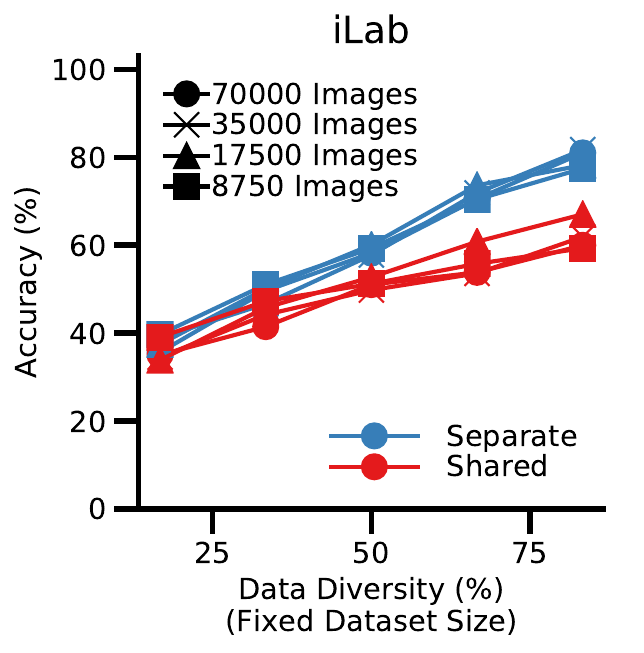}
\end{tabular}
 \caption{\emph{Generalization to OOD combinations as number of training images is varied.} For both iLab and Biased-Cars dataset, \textit{Separate} architecture outperforms the \textit{Shared} architecture trained with the same number of images.}
 \label{fig:data_control}
\end{figure*}

To ensure that our findings are not a function of the amount of training data, we present the results for different number of images for the \emph{Biased-Cars}  and the iLab dataset in Fig.~\ref{fig:data_control}. As can be seen in both these datasets, across a different number of images the \textit{Separate} architecture substantially outperforms the \textit{Shared} one at generalizing to OOD category-viewpoint combinations.

\FloatBarrier

\subsection{Results on new categories and viewpoints} 

\label{app:DifferentBias}
\begin{figure*}[t!]
\centering
\begin{tabular}{@{\hspace{0.cm}}c@{\hspace{-0.1cm}}c}
     \includegraphics[width=0.44\linewidth]{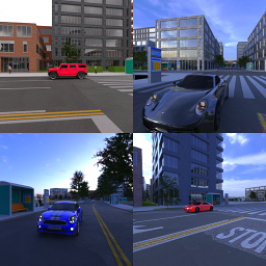}
&
     \includegraphics[width=0.44\linewidth]{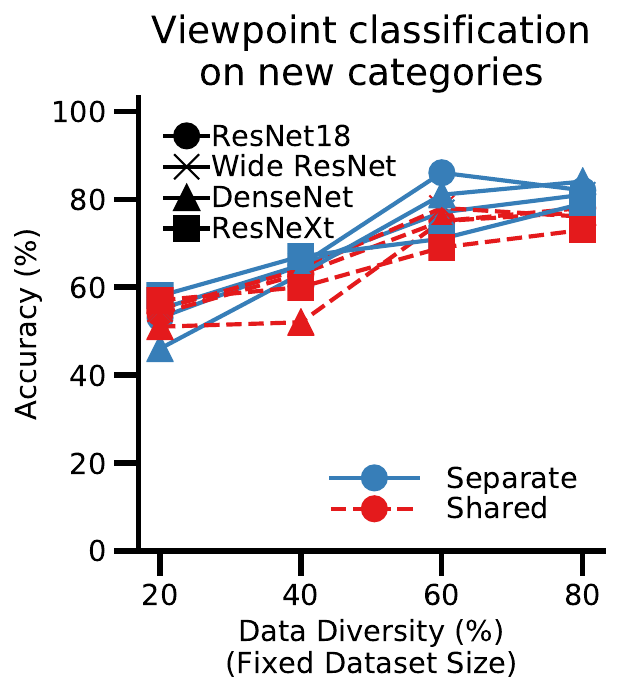}\\
     {(a)} & {(b)} \\
 \includegraphics[width=0.44\linewidth]{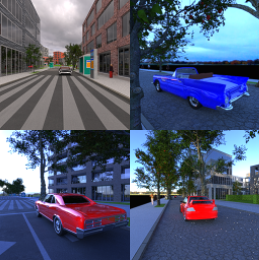}
&  \includegraphics[width=0.44\linewidth]{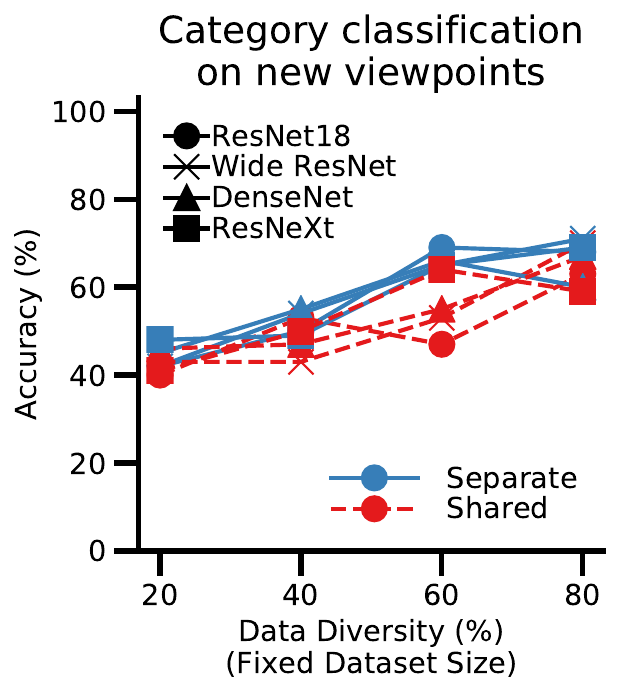}\\
 {(c)} & {(d)}  \vspace{-2ex}
\end{tabular}
\caption{\emph{Controlling category and viewpoint separately.}
 (a) Images of 4 new car categories, (b) Viewpoint classification accuracy for the 4 new car categories, (c) Images of new viewpoints, (d) Car category recognition accuracy for the new viewpoints.}
 \label{fig:rebuttal_exp}
\end{figure*}

We also evaluated our trained CNNs on classifying the viewpoint of 4 new car categories (Fig.~\ref{fig:rebuttal_exp}a). Analogously, we also evaluated category classification in new viewpoints (side-to-back of car as in Fig.~\ref{fig:rebuttal_exp}c, instead of the front-to-side shown in training). As shown in Fig.~\ref{fig:rebuttal_exp}b and d, these results confirm that our conclusions also apply to new car categories and new viewpoints: generalization increases with more data diversity and \textit{Separate} architecture.

\FloatBarrier

\subsection{Task training order}

\label{app:TaskOrder}

\begin{figure*}[!t]
\begin{tabular}{@{\hspace{0.1cm}}c@{\hspace{0.1cm}}c}
\centering\includegraphics[width=0.5\linewidth]{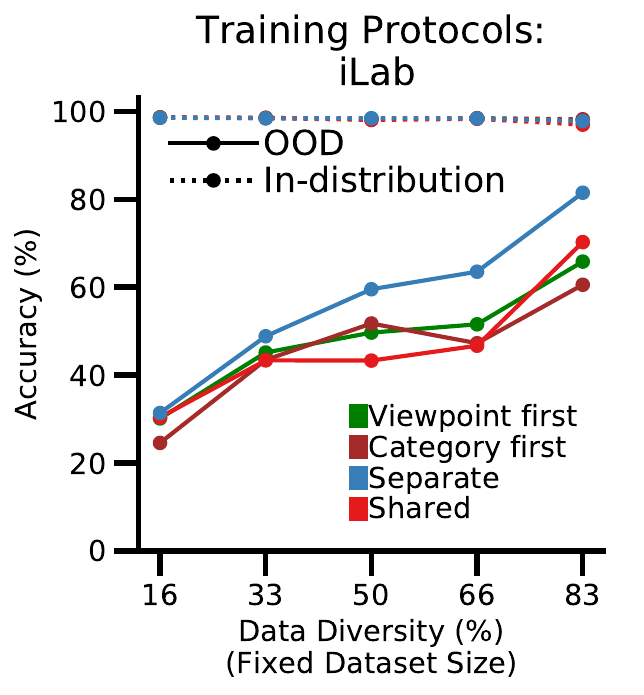}&
     \includegraphics[width=0.5\linewidth]{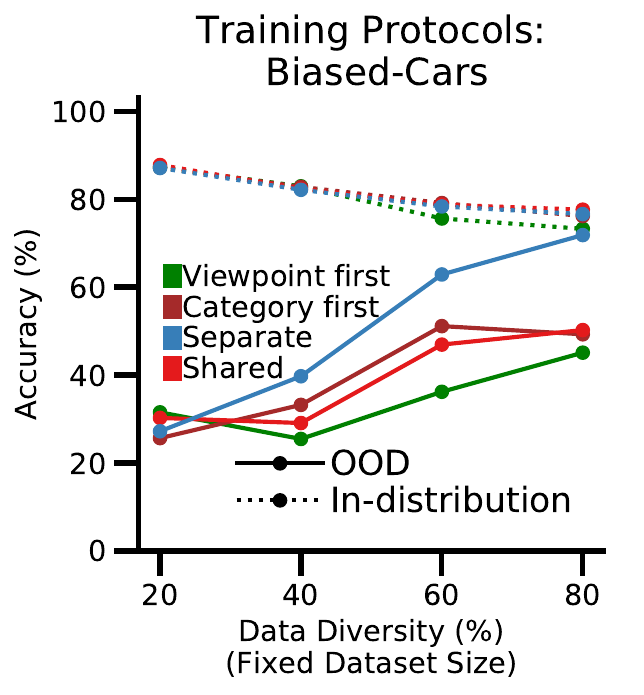}
\end{tabular}
\caption{\emph{Generalization performance for additional training protocols, besides Separate and Shared protocols, for the ResNet-18 backbone.} The geometric mean of category recognition accuracy and viewpoint estimation accuracy is reported for OOD combinations as the number  of in-distribution combinations  is increased. For these, we start by training the shared network on only one task first, \ie \textit{Viewpoint  first} or \textit{Category first}. We then train the second task starting from these learned features from the first task. We present their comparison with our \textit{Shared} and \textit{Separate} training protocols presented in the main paper. (a) Accuracy for the iLab dataset, (b) Accuracy for the Biased-Cars dataset.}
\label{fig:training_protocol}
\end{figure*}

We present results on the impact of the order in which networks are trained on category and viewpoint classification. 
Our networks contain three components: (i) shared layers, (ii) category branch and (iii) viewpoint branch. Here we start by training on one task first, say Category recognition. We then train the other task, \ie Viewpoint classification starting from these features learned from the first task. We call this the \textit{Category first} protocol. The \textit{Viewpoint first} protocol is defined analogously by starting with viewpoint classification first, and then training for category recognition. Results for these are provided in Fig. \ref{fig:training_protocol}.\\

As can be seen, our findings are consistent with these new protocols as well. The \textit{Separate} architecture outperforms the \textit{Shared} architectures independent of the training protocol. Furthermore, all architectures get better with OOD combinations as in-distribution combinations are increased.
\FloatBarrier

\subsection{Results on additional datasets: UIUC 3D and MNIST-Rotation}
\label{app:AdditionalDatasets}
\begin{figure*}[!t]
\begin{tabular}{@{\hspace{0.1cm}}c@{\hspace{0.1cm}}c}
\centering\includegraphics[width=0.5\linewidth]{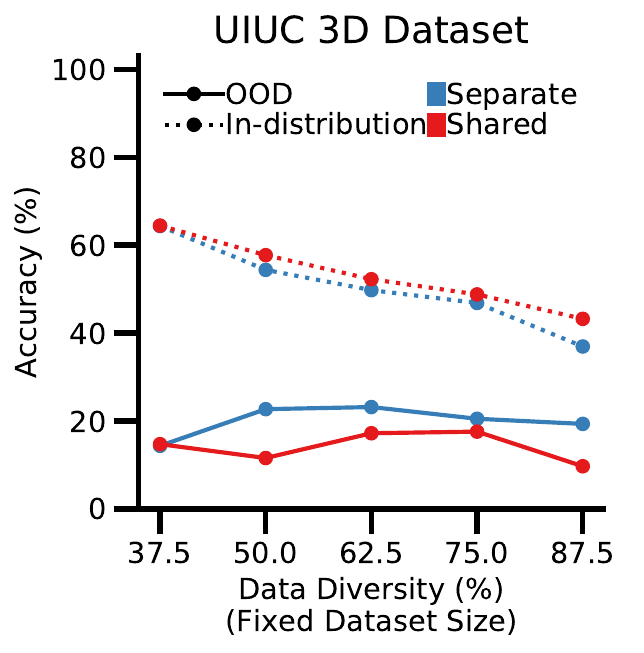}&
     \includegraphics[width=0.5\linewidth]{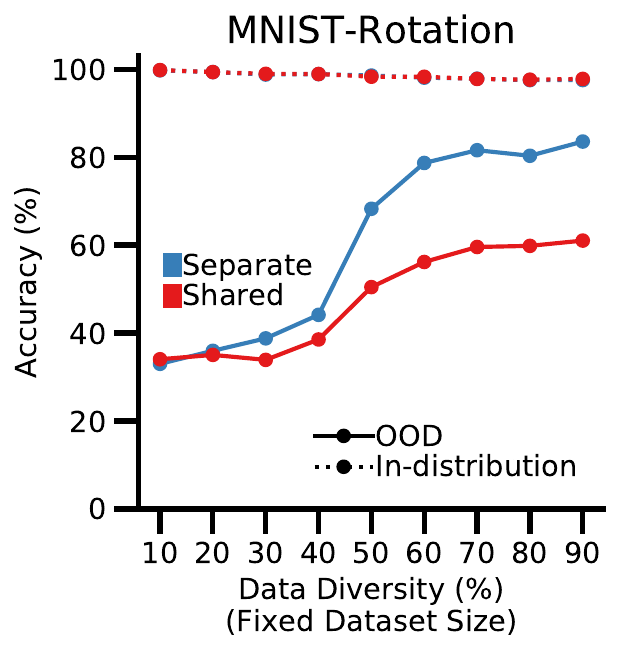}
\end{tabular}
\caption{\emph{Generalization performance for additional datasets for ResNet-18 backbone.} The geometric mean of category recognition accuracy and viewpoint estimation accuracy is reported for OOD combinations as the number  of in-distribution combinations is increased. (a) Accuracy for the UIUC 3D dataset, (b) Accuracy for the MNIST-Rotation dataset. Due to the small size of the UIUC dataset there is poor generalization - there are only 1700 train set for 37.5\% in-distribution combinations (which is kept constant as the number of in-distribution combinations increases). This leads to lesser generalization, but our findings still hold true - (1) Increasing in-distribution combinations improves performance on OOD data, and (2) \textit{Separate} architectures outperform \textit{Shared} ones on OOD combinations.}
\label{fig:uiuc_3d_mnist_rotation}
\end{figure*}

Going beyond the four datasets presented in the main paper, we replicate our analysis on two additional datasets as a confirmatory experiment: (1) the UIUC 3D Dataset, and (2) the MNIST-Rotation dataset. As can be seen from Figs.~\ref{fig:uiuc_3d_mnist_rotation} (a) and (b), our findings are consistent across these additional datasets as well - \textit{Separate} outperforms the \textit{Shared}, and all architectures get better at OOD combinations as in-distribution combinations are increased. \\

\textbf{Small size of UIUC 3D dataset:} It is important to note that the small size of the UIUC 3D dataset makes it difficult to adapt it for training with biased in-distribution combinations. We picked 8 of the total 10 object categories (to ensure symmetry between tasks as explained in the paper), which amounts to 5,400 images in total across 64 category-viewpoint combinations. Thus, there are only 1700 training images for the 24 in-distribution combinations case, which is kept constant as in-distribution combinations are increased. In contrast, the other natural image dataset used in this paper, the iLab dataset contains 70,000 training images for 6 categories and viewpoints each. Due to this the generalization performance is slightly low, however the findings are still consistent as reported above. As an additional control, we also tried using all available 4500 images for the 87.5\% seen case (\ie all images other than the OOD test set) - generalization numbers were still low overall, but trends were preserved.
\FloatBarrier

\subsection{Results on group equivariant architectures}
\label{app:AdditionalArch}
\begin{figure*}[!t]
\begin{tabular}{@{\hspace{0.0cm}}c@{\hspace{0.1cm}}c}
\centering\includegraphics[width=0.5\linewidth]{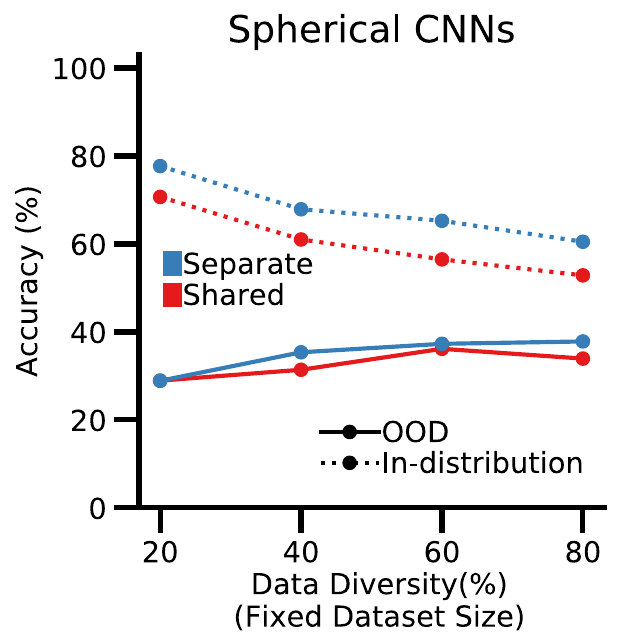}&
     \includegraphics[width=0.5\linewidth]{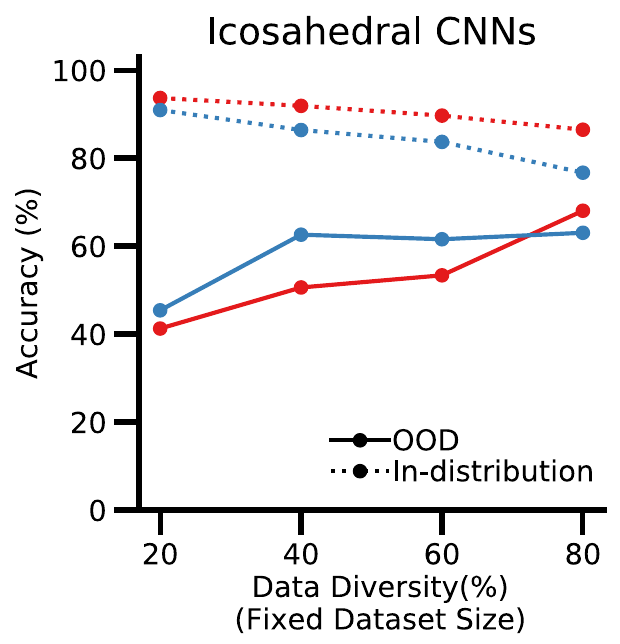}
\end{tabular}
 \caption{\emph{Generalization performance for different group equivariant architectures as in-distribution combinations are increased for MNIST-Rotation dataset.} The geometric mean of category recognition accuracy and viewpoint estimation accuracy is reported for OOD combinations as the number  of in-distribution combinations  is increased. (a) Accuracy of \textit{Separate} and \textit{Shared} architectures using a Spherical CNN~\citep{cohen2018spherical} as backbone, (b) Accuracy using an Icosahedral CNN~\citep{cohen2019gauge} as backbone.}
\label{fig:gcnns}
\end{figure*}

Group and gauge equivariant CNNs have recently emerged as an alternative to standard CNNs which theoretically offer better viewpoint invariance. While these architectures~\citep{cohen2018spherical, cohen2019gauge} are yet to be adapted to more complex datasets like ImageNet, they have shown great results on simpler image datasets like MNIST-Rotation. Here, we present results with two such architectures in Fig.~\ref{fig:gcnns}. As can be seen, our findings also extend to these architectures -  \textit{Separate} outperforms the \textit{Shared} independent of the training protocol, and all architectures get better at OOD combinations as in-distribution combinations are increased. This suggests our findings extend beyond standard CNNs. We believe that a detailed comparison between GCNNs and standard CNNs with respect to generalization to OOD combinations would be an interesting starting point for future work.

\newpage
\section{Additional Experiments for ``How Do CNNs Generalize to OOD Combinations?''}

\subsection{Ratio of Specialized Neurons}
\label{app:SpecializationAdditionalDatasets}

\begin{figure*}[!t]
\begin{tabular}{@{\hspace{-0.cm}}c@{\hspace{-0.1cm}}c@{\hspace{-0.1cm}}c@{\hspace{-0.1cm}}c}
     \includegraphics[width=0.255\linewidth]{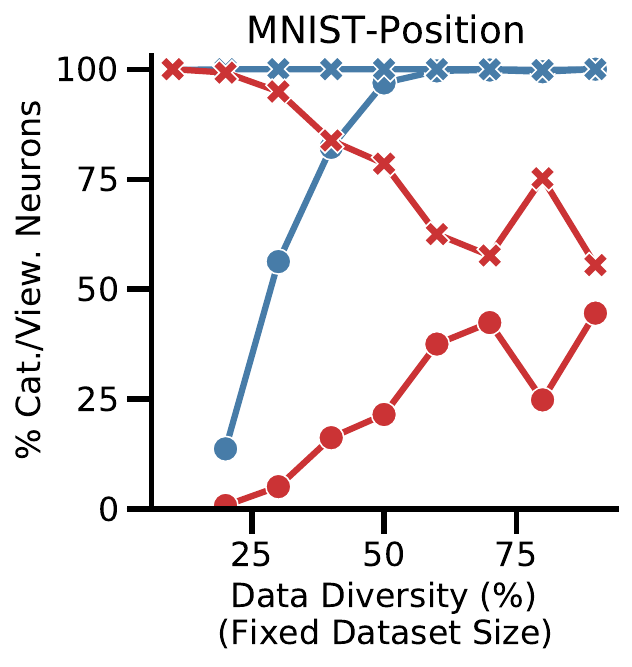}
&
     \includegraphics[width=0.255\linewidth]{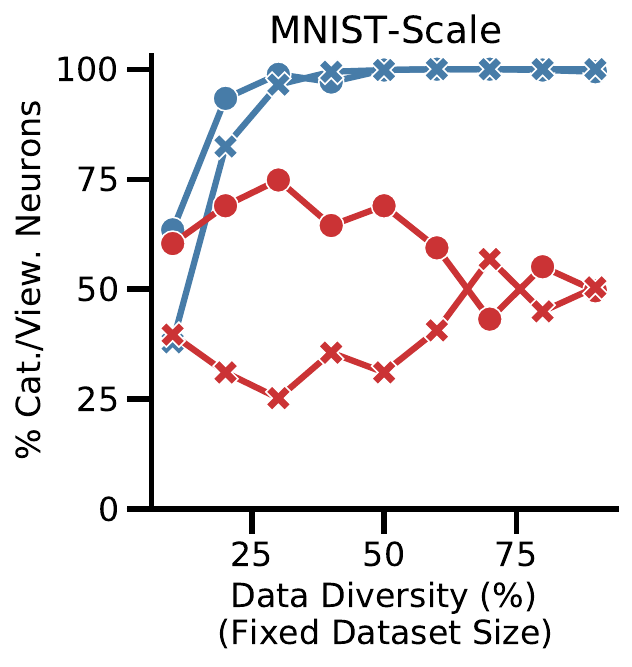} 
&  \includegraphics[width=0.255\linewidth]{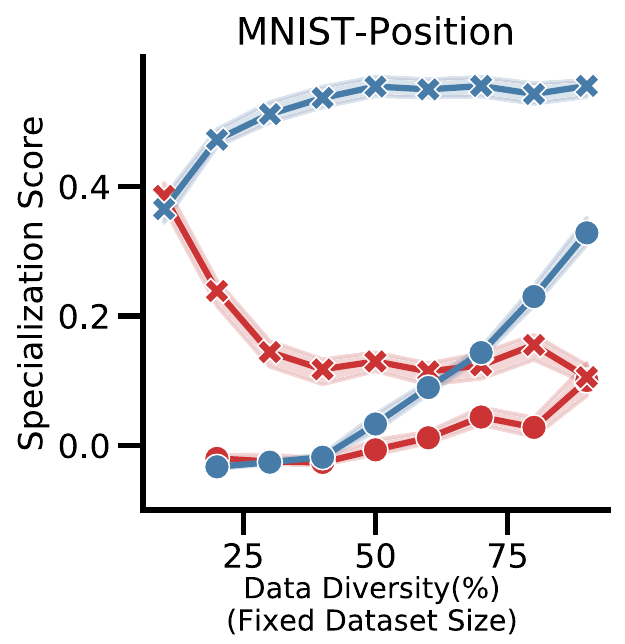} &  \includegraphics[width=0.255\linewidth]{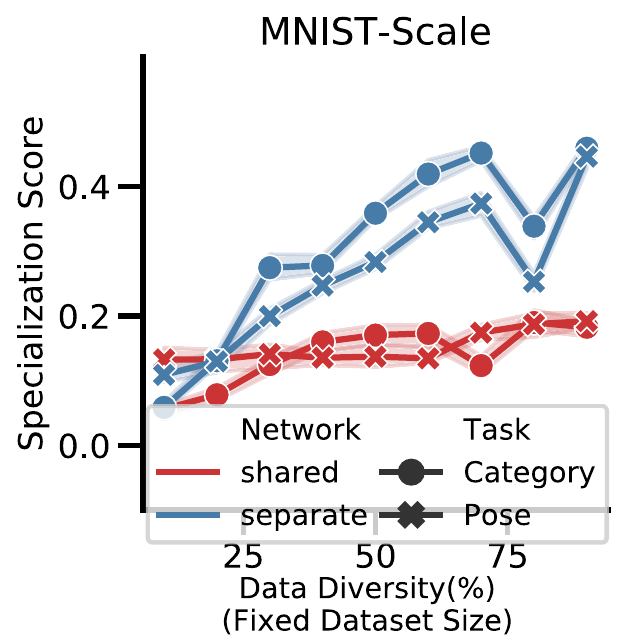}   \\
(a) & (b) & (c)& (d) \\
\end{tabular}
\caption{\emph{Neuron specialization in MNIST-Position and MNIST-Scale datasets}. (a) and (b) Percentage of neurons in the final convolutional layer of ResNet-18 that are specialized to category and viewpoint, for MNIST-Position and MNIST-Scale datasets, respectively. (c) and (d) Median of the specialization scores of neurons in the final convolutional layer of ResNet-18 \emph{Separate} and \emph{Shared} architectures, for category and viewpoint classification tasks, respectively.}
\label{fig:generalSation_datasets_tasks}
\end{figure*}

In the main paper, we have presented the ratio of specialized neurons for the iLab and \textit{Biased-Cars} dataset. Here, we also provide these for the MNIST-Position  and MNIST-Scale datasets. As can be seen, our findings are consistent across these datasets as well. Figs.~\ref{fig:generalSation_datasets_tasks}a  and b show that neurons in the final convolutional layer specialize to become either category or viewpoint neurons as more category-viewpoint combinations are shown. Category and viewpoint branches of the \textit{Separate} architecture become completely specialized to category and viewpoint, respectively. In the \textit{Shared} architecture, both kinds of neurons emerge in roughly equal numbers. 

We also observe that for a small number of in-distribution combinations, the ratio of neurons specialized for category or viewpoint classification may be impacted by the relative difficulty of these two tasks. 
We observe that when the accuracy is higher for category classification (shown in Fig.~\ref{fig:accuracy_tasks}), a higher fraction of neurons becomes specialized for category, as observed for the iLab and MNIST-scale datasets. Similarly, when accuracy for viewpoint classification is higher, a greater fraction of neurons becomes specialized for viewpoint, as observed in  MNIST-Position.

\subsection{Specialization score for additional datasets}
\label{app:specializationiLab}
\begin{figure*}[!h]
\begin{tabular}{@{\hspace{-0.cm}}c@{\hspace{-0.1cm}}c@{\hspace{-0.1cm}}c@{\hspace{-0.1cm}}c}

\includegraphics[width=0.255\linewidth]{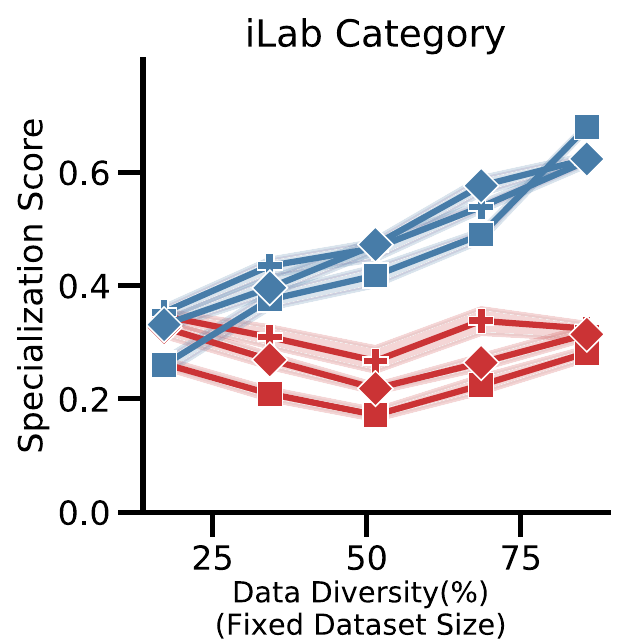}
&
\includegraphics[width=0.255\linewidth]{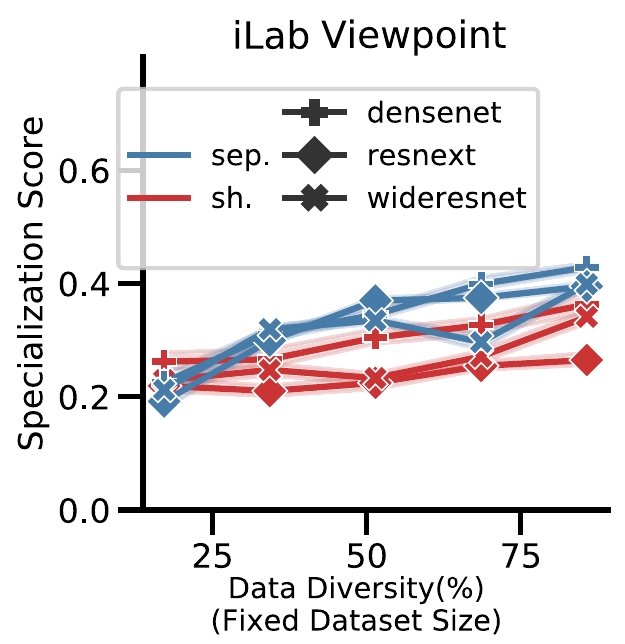}
& 
\includegraphics[width=0.255\linewidth]{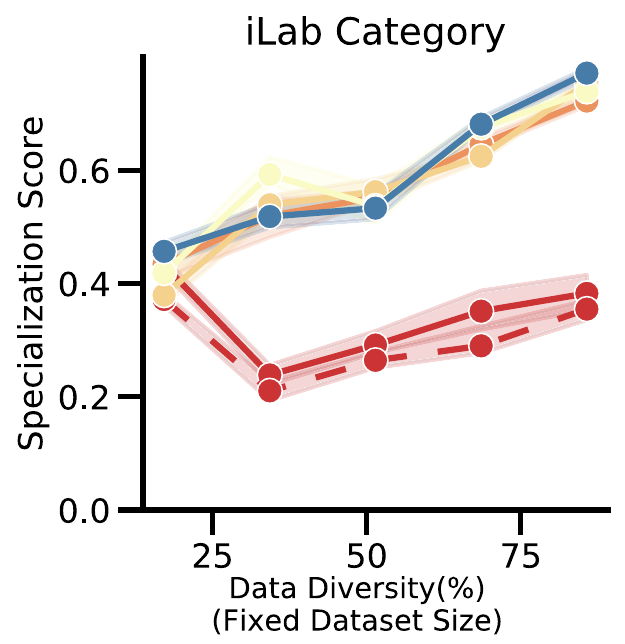}
& \includegraphics[width=0.255\linewidth]{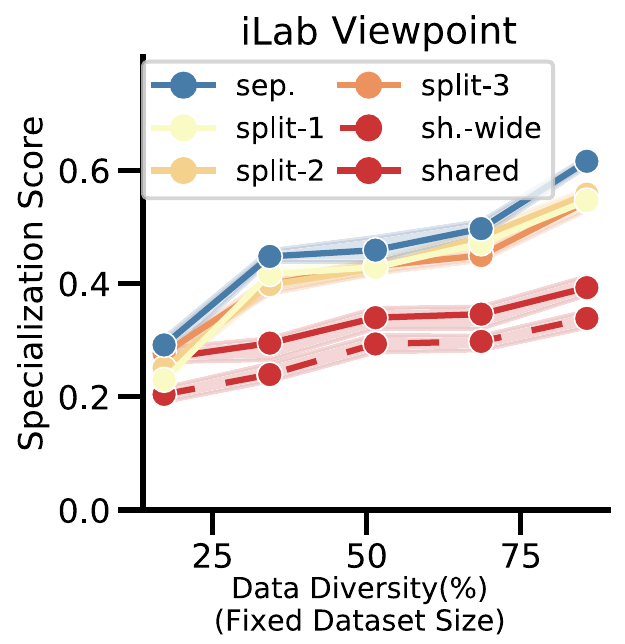}\\
(a) & (b) & (c)& (d) \\
\end{tabular}
\caption{\emph{Neuron specialization (selectivity to category and invariance to viewpoint, and vice versa) in the iLab dataset.} 
(a) and (b) Median of the specialization score among neurons ($\Gamma^k$) in network architectures, other than ResNet-18, \emph{separate} and \emph{shared}, for category and viewpoint classification tasks, respectively.  Confidence intervals (95\%) are displayed in low opacity.
(c) and (d) Median of the specialization score among neurons in ResNet-18 \emph{Separate} and \emph{Shared} with splits made at different blocks of the network, for category and viewpoint classification tasks, respectively. Similar results for the \textit{Biased-Cars} dataset are provided in the main paper.}
\label{fig:specialization_scores_iLab}
\end{figure*}

Figs.~\ref{fig:generalSation_datasets_tasks}c and d show that as the number of in-distribution combinations are increased, there is a steady increase in the specialization score for both MNIST-Position and MNIST-Scale. 
In Fig.~\ref{fig:specialization_scores_iLab}, we show that the selectivity score results are also consistent in iLab for different backbones and split architectures.

\subsection{Invariance and Selectivity Scores}
\label{app:InvarianceAndSelectivity}
\begin{figure*}[!t]
\begin{tabular}{@{\hspace{-0.cm}}c@{\hspace{-0.1cm}}c@{\hspace{-0.1cm}}c@{\hspace{-0.1cm}}c}

\includegraphics[width=0.255\linewidth]{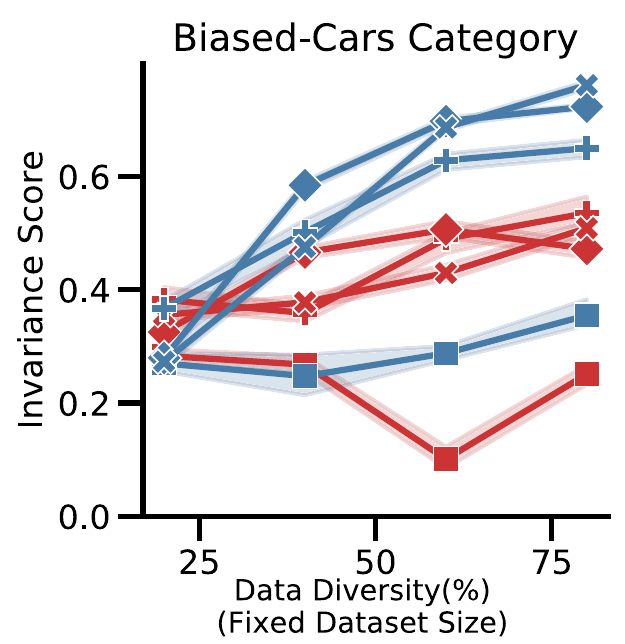}
&
\includegraphics[width=0.255\linewidth]{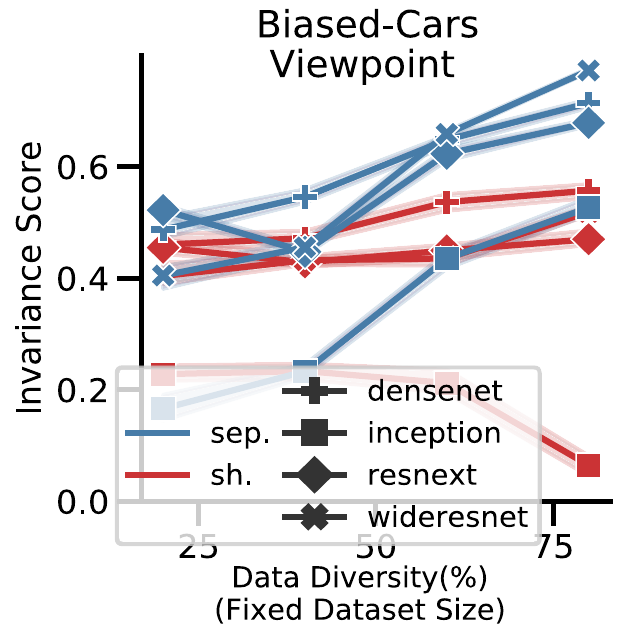}
& 
\includegraphics[width=0.255\linewidth]{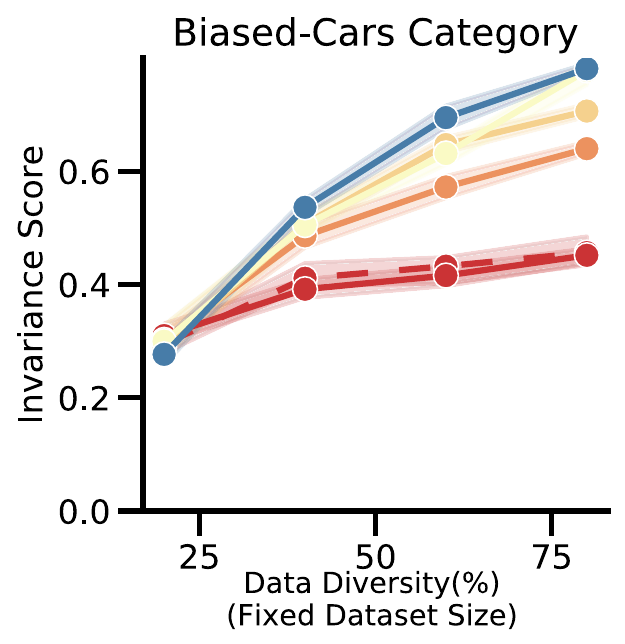}
& \includegraphics[width=0.255\linewidth]{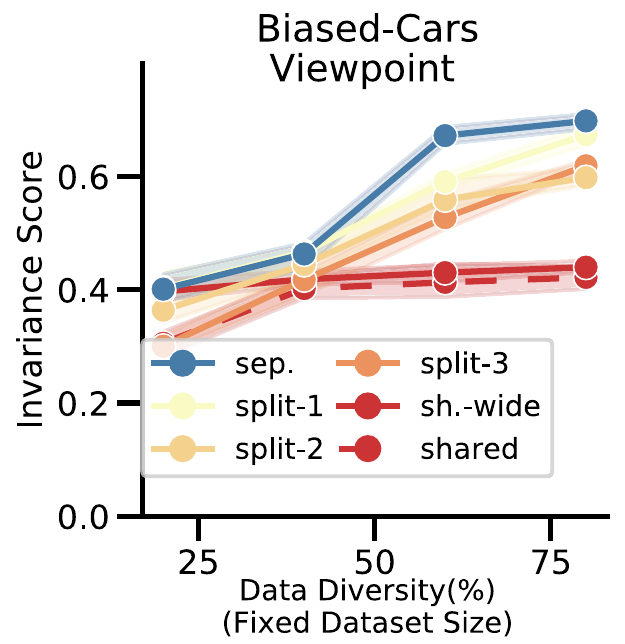}\\
(a) & (b) & (c)& (d) \\
\end{tabular}
\caption{\emph{Invariance scores in the Biased-Cars dataset.} 
(a) and (b) Median of the invariance score among neurons in network architectures, other than ResNet-18, \emph{separate} and \emph{shared}, for category and viewpoint recognition tasks, respectively. Confidence intervals (95\%) are displayed in low opacity.
(c) and (d) Median of the invariance score among neurons in ResNet-18 \emph{Separate} and \emph{Shared} with splits made at different blocks of the network, for category and viewpoint recognition tasks, respectively. }
\label{fig:specialization_scores_invariance}
\end{figure*}

\begin{figure*}[!t]
\begin{tabular}{@{\hspace{-0.cm}}c@{\hspace{-0.1cm}}c@{\hspace{-0.1cm}}c@{\hspace{-0.1cm}}c}

\includegraphics[width=0.255\linewidth]{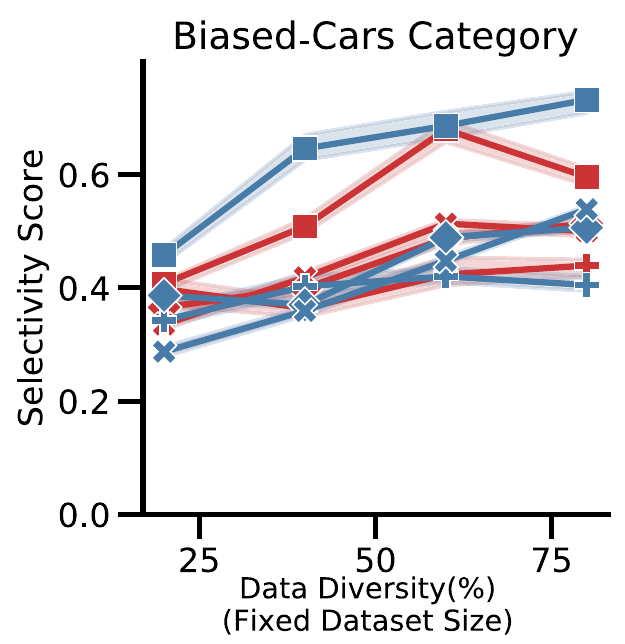}
&
\includegraphics[width=0.255\linewidth]{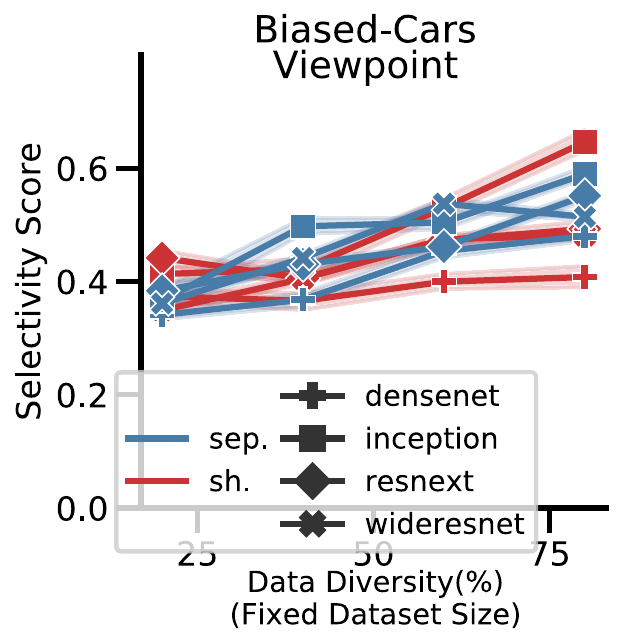}
& 
\includegraphics[width=0.255\linewidth]{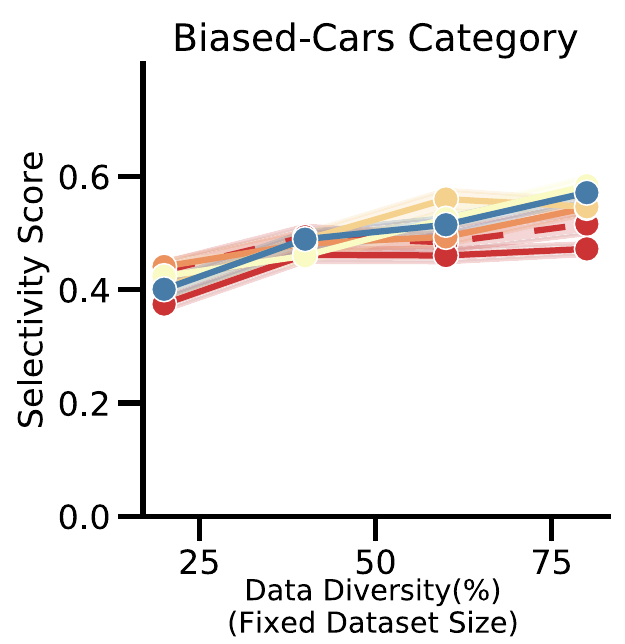}
& \includegraphics[width=0.255\linewidth]{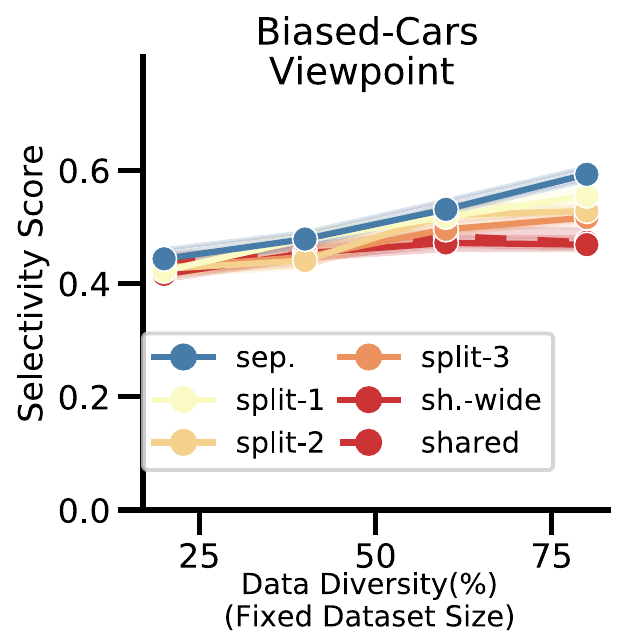}\\
(a) & (b) & (c)& (d) \\
\end{tabular}
\caption{\emph{Selectivity scores in the Biased-Cars dataset.} 
(a) and (b) Median of the selectivity score among neurons in network architectures, other than ResNet-18, \emph{separate} and \emph{shared}, for category and viewpoint recognition tasks, respectively. Confidence intervals (95\%) are displayed in low opacity.
(c) and (d) Median of the selectivity score among neurons in ResNet-18 \emph{Separate} and \emph{Shared} with splits made at different blocks of the network, for category and viewpoint recognition tasks, respectively. }
\label{fig:specialization_scores_selectivity}
\end{figure*}

In Fig.~\ref{fig:specialization_scores_invariance} and~\ref{fig:specialization_scores_selectivity}, we show the invariance and selectivity scores separately for the \emph{Biased-Cars} dataset. In both cases, the trends follow what we observed for the specialization score, though the differences are much more pronounced in terms of invariance rather than selectivity.

\subsection{Specialization Score per Layer}
\label{app:SpecializationBuilds}
\begin{figure*}[!t]
\begin{tabular}{@{\hspace{-0.cm}}c@{\hspace{-0.1cm}}c@{\hspace{-0.1cm}}c@{\hspace{-0.1cm}}c}

\includegraphics[width=0.255\linewidth]{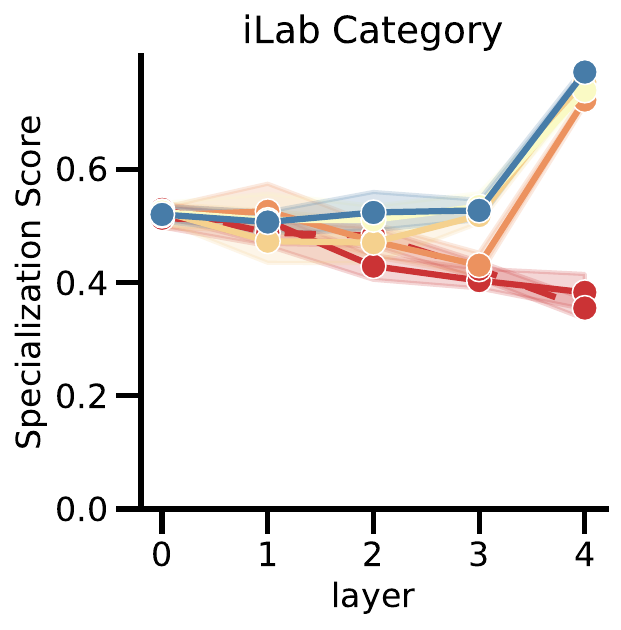}
&
\includegraphics[width=0.255\linewidth]{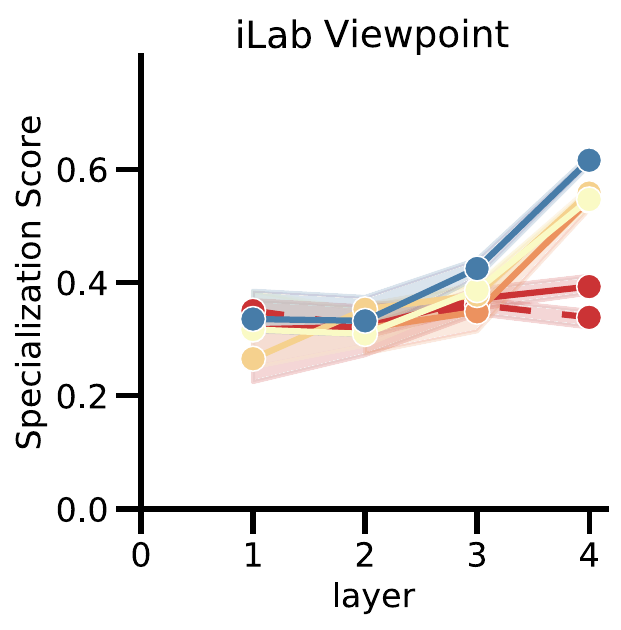}
& 
\includegraphics[width=0.255\linewidth]{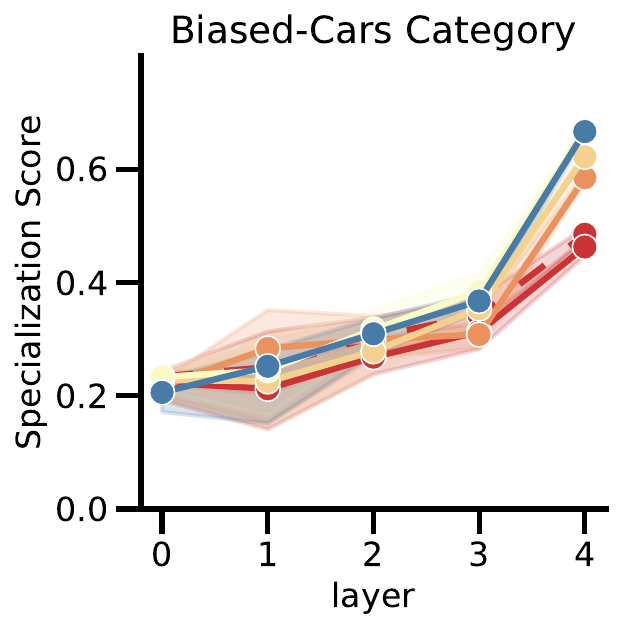}
& \includegraphics[width=0.255\linewidth]{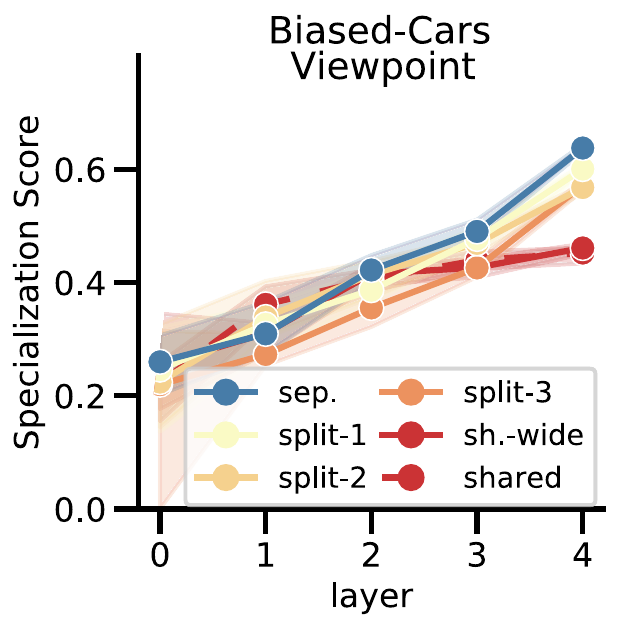}\\
(a) & (b) & (c)& (d) \\
\end{tabular}
\caption{\emph{Specialization Score Per Layer for $30$ \emph{seen} category-viewpoint Combinations for iLab, and $20$ \emph{seen} category-viewpoint Combinations for the \textit{Biased-Cars} dataset.} 
(a) and (b) Median of the specialization score among neurons in ResNet-18 \emph{Separate} and \emph{Shared} with splits made at different blocks of the network, for category and viewpoint classification tasks, respectively. (c) and (d) Same as (a) and (b) for \emph{Biased-Cars} dataset. }
\label{fig:specialization_scores_layers}
\end{figure*}

In Fig.~\ref{fig:specialization_scores_layers}, we show the specialization score in each layer. We can see that it builds up across layers, and this is more pronounced for \emph{Separate} architectures than for \emph{Shared}.

\newpage
\section{Limitations}
\label{sec:suplLimits}

In this paper we have only considered rigid objects, while general object recognition often involves deformable and articulated object categories including humans and other animals.For such objects, parts may appear in various configurations for the same viewpoint. One way to analyze this more complex scenario would be to extend our experiments to study combinations of configurations, viewpoints and categories. Furthermore, this analysis may also be extended to study the impact of object symmetries, which would alter the effective number of visually distinct object viewpoints.


Also, we have considered selectivity and invariance of individual neurons as a model for understanding generalization to OOD combinations. This model is limited in several ways as it only considers the properties of individual neurons, and assumes that selectivity to one single category (or viewpoint) is needed alongside invariance to viewpoint (or category) to achieve generalization. There could be other ways to achieve generalization not taken into account by the model. Also, the evidence presented here is correlational and based on the average neural activity for a set of images. Nonetheless, the model has been shown to be useful to explain in simple and intuitive terms why the \emph{Separate} architecture outperforms the \emph{Shared} one, and how these generalize as more category-viewpoint combinations are seen.

\end{document}


\maketitle
In this document we provide additional details and experiments which were not presented in the main submission for the sake of brevity. This document follows the same order as the main paper.

\fi